\documentclass[pdflatex,iicol,sn-apa]{sn-jnl}

\usepackage{amsmath}
\usepackage{amssymb}
\usepackage{pifont}
\usepackage{graphicx}
\usepackage{xcolor}
\usepackage{multirow}
\usepackage{natbib}
\usepackage{bbm}
\usepackage{latexsym}
\usepackage{url}            

\newcommand{\cmark}{\ding{51}}%
\newcommand{\xmark}{\ding{55}}%

\newcommand{\sgap}{\vspace*{3mm}}

\newcommand{\noi}{\noindent}

\newcommand{\bfa}{\mathbf{a}}
\newcommand{\bfb}{\mathbf{b}}
\newcommand{\bfc}{\mathbf{c}}
\newcommand{\bfd}{\mathbf{d}}
\newcommand{\bff}{\mathbf{f}}

\newcommand{\bfo}{\mathbf{o}}
\newcommand{\bfp}{\mathbf{p}}
\newcommand{\bfr}{\mathbf{r}}
\newcommand{\bfs}{\mathbf{s}}
\newcommand{\bft}{\mathbf{t}}
\newcommand{\bfw}{\mathbf{w}}
\newcommand{\bfx}{\mathbf{x}}

\newcommand{\bfz}{\mathbf{z}}

\newcommand{\cut}[1]{}

\mathchardef\mhyphen="2D

\begin{document}

\title[Structured Generative Models for Scene Understanding]{Structured Generative Models for Scene Understanding}

\author{\fnm{Christopher K. I.} \sur{Williams}}\email{c.k.i.williams@ed.ac.uk}

\affil{\orgdiv{School of Informatics}, \orgname{University of
    Edinburgh}, \orgaddress{\street{10 Crichton Street},
    \city{Edinburgh}, \postcode{EH8 9AB}, \country{United Kingdom}}}

\abstract{
This position paper argues for the use of \emph{structured generative
models} (SGMs) for the understanding of static scenes. This
requires the reconstruction of a 3D scene from an input image
(or a set of multi-view images), whereby the contents of the
image(s) are causally explained in terms of models of instantiated
objects, each with their own type, shape, appearance and pose, along
with global variables like scene lighting and camera parameters. This
approach also requires scene models which account for the
co-occurrences and inter-relationships of objects in a scene. The SGM
approach has the merits that it is compositional and generative, which
lead to interpretability and editability. \\\\
To pursue the SGM agenda, we need models for objects and scenes, and
approaches to carry out inference.  We first review models for
objects, which include ``things'' (object categories that have a well
defined shape), and ``stuff'' (categories which have amorphous spatial
extent). We then move on to review \emph{scene models} which describe
the inter-relationships of objects.  Perhaps the most challenging
problem for SGMs is \emph{inference} of the objects, lighting and
camera parameters, and scene inter-relationships from input consisting
of a single or multiple images.  We conclude with a discussion of issues
that need addressing  to advance the SGM agenda.
}

\keywords{structured generative models, generative models,
compositionality, scene understanding}

\maketitle

\section{Introduction}

The goal of this position paper is to promote the use of
\emph{structured generative models} (SGMs) for the understanding
of static scenes.
These models are situated in the classical framework for computer
vision whereby a 3D scene is \emph{reconstructed} from one or more
input images.  In this case the contents of the image are causally
explained in terms of models of instantiated objects, each with their
own type, shape, appearance and pose, along with global variables like
scene lighting and camera parameters.  This approach also
requires \emph{scene models} which account for the co-occurrences and
inter-relationships of objects in a scene.  Because such models
can \emph{generate} (or reconstruct, or explain) the scene, and
because they are \emph{structured} (i.e.\ they are composed of
multiple objects and their relationships), we term them
\emph{structured generative models}.

This reconstructive framework is also known as \emph{analysis-by-synthesis}
\citep{grenander-78}, or \emph{vision-as-inverse-graphics (VIG)}
\citep{kulkarni-whitney-kohli-tenenbaum-15,moreno-williams-nash-kohli-16}.
It can be traced back to the early days of
``blocks world'' research in the 1960s \citep{roberts-63}. Other early work
in this vein includes the VISIONS system
of \citet{hanson-riseman-visions-78}, 
and the system of \citet{ohta-kanade-sakai-78} for outdoor scene
understanding. 
For example, the VISIONS system used various levels 
of analysis (e.g., objects, surfaces), and mappings 
between the image-specific parse and generic knowledge about scenes.

Alternatives to the VIG framework are either \emph{discriminative
approaches}, predicting some target quantity or quantities given input
image(s), or
\emph{unstructured generative models}. Discriminative approaches
are typically applied to solve \emph{specific
tasks} such as object detection or semantic segmentation, 
which are usually specified in \emph{image
space}. These are usually set up as supervised learning tasks, thus
requiring annotated data. Currently deep neural networks (DNNs)
are the dominant method-of-choice for such tasks. 
DNNs are often highly accurate, but as
discriminatively-trained models they can sometimes fail badly,
producing absurd mistakes,\footnote{The phrase ``absurd mistakes'' is
borrowed from Daniel Kahneman's talk at the NeurIPS conference in
December 2021.} but with no effective indication of unreliability. One
example of this is performance failures on adversarial examples (see
e.g.,
\citealt*{szegedy-zaremba-sutskever-bruna-erhan-goodfellow-fergus-13}).
Also as the discriminative models are trained on specific datasets,
they can often perform poorly when faced with the same task but on a
novel dataset with different statistics (distribution shift).
The focus on the evaluation of specific tasks means that the predictions
from multiple tasks are not required to create a coherent understanding of the
input image in terms of the 3D world; this point is made, e.g.,
  by \citet{zamir-sax-consistency-20}.

With \emph{unstructured generative models}, images are generated from
a single set of latent variables, without explicit modelling of
objects and their interactions.  An example is the work
of \citet{radford-metz-chintala-16} where images of bedroom scenes
generated by generative adversarial networks (GANs). Here there is a
single latent vector representation for the whole scene, which is not
disentangled across objects. Similar vectorial representations
are used in more recent generative models of images, such as
DALL-E
\citep{ramesh-dalle-21} and Gemini \citep{gemini-23} which
are based on transformer decoders, and diffusion models like that
due to \citet{rombach-stable-diffusion-22}, and  GLIDE \citep{nichol-glide-22}.
(These recent models can be made conditional on vectorial
representations of text or other input modalities.)
Despite impressive-looking output from these models, the unstructured
nature of the latent representation means that it can be hard 
to edit it to make specific changes
in the scene (e.g., to change the colour of a bedspread), as the
representation is not interpretable.

To be clear, we are not arguing against the use of deep neural
networks in computer vision. However, for structured generative models
DNNs can be used for specific modelling and inference tasks (as we
will see below), rather than as one big black box.
In contrast to a supervised DNN, a SGM provides a coherent scene
representation that can subserve multiple tasks, including
novel ones. The use of a scene model allows top-down information
flows (e.g., about the type of scene) into tasks such as the detection
and segmentation of objects.  The \emph{structured} aspect means that
the representation is interpretable, and this allows editing of the
scene, e.g., to add objects, or to change the properties of objects in
the scene. The scene model can be made \emph{compositional}, being
composed hierarchically from more elementary substructures.
As \citet{yuille-liu-21} argue, this should allow SGMs to handle the
combinatorial explosion of possible images without needing exponential
amounts of training data. These issues are expanded upon below, and
particularly in sec.\ \ref{sec:pros_cons}.

\begin{figure*}
  \begin{tabular}{ccc}
    \includegraphics[height=1.5in]{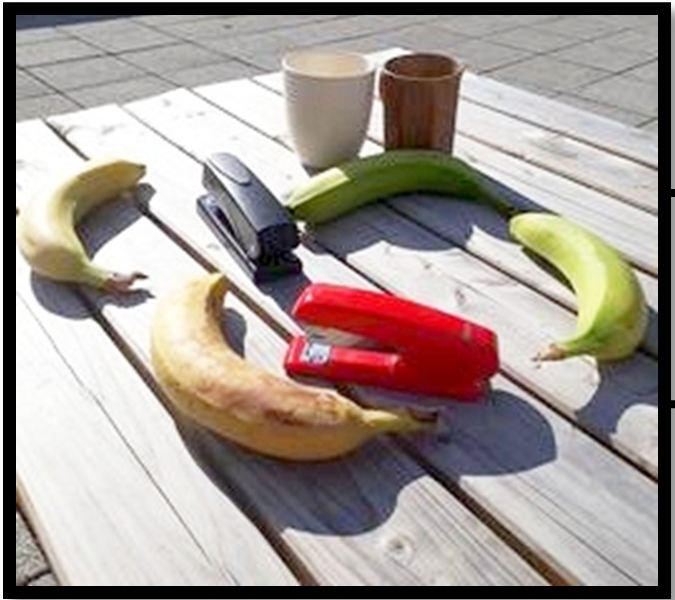} & 
    \includegraphics[height=1.9in]{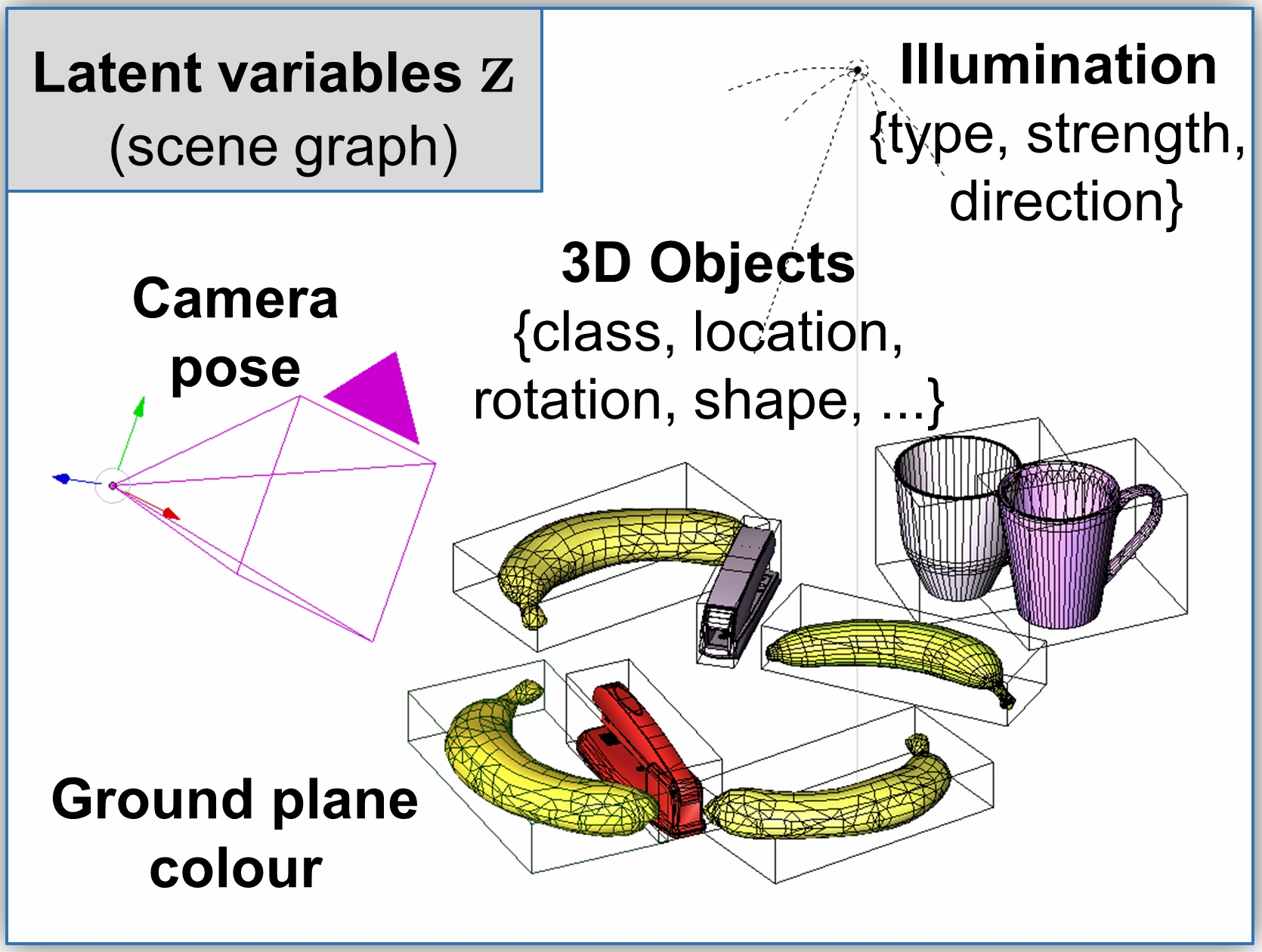} & 
    \includegraphics[height=1.5in]{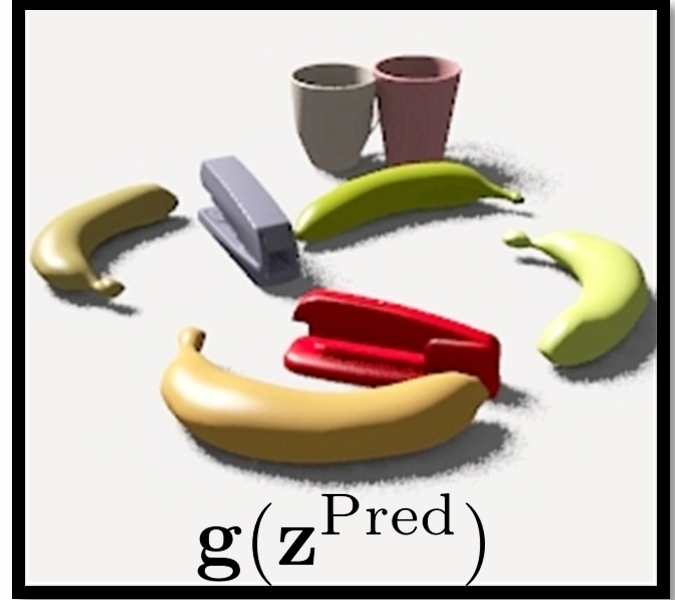} \\
    input &  latent variables & reconstruction
  \end{tabular}    
  \caption{The input image (left) is explained in terms of 3D objects,
    the camera pose and illumination, to produce the reconstructed
    image (right). Images from \citet{romaszko-williams-winn-20}.
 \label{fig:lukasz}}
\end{figure*}

\begin{figure*}
\begin{tabular}{cccc}
   \includegraphics[height=0.67in]{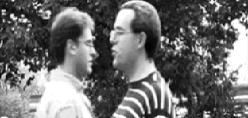} & 
   \includegraphics[height=0.67in]{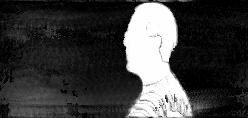} &
   \includegraphics[height=0.67in]{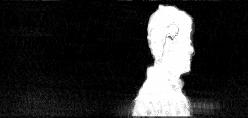} &    \\
   \includegraphics[height=0.67in]{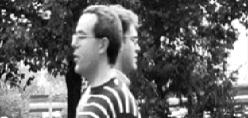} & 
   \includegraphics[height=0.67in]{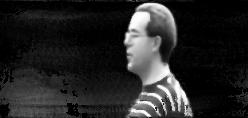} &
   \includegraphics[height=0.67in]{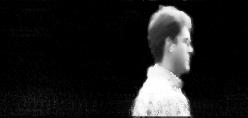} &   
   \includegraphics[height=0.67in]{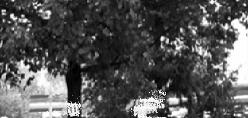}    
\end{tabular}    
\caption{The left most panel shows two frames from a video
    of two people walking past each other against a background.
    The second panel shows the
    mask (top) and appearance of the first sprite learned. The third
    panel shows the same thing for the second sprite. The rightmost
    panel shows the learned background. Images from \citet{williams-titsias-04}.
 \label{fig:fj}}
\end{figure*}

Fig.\ \ref{fig:lukasz} shows how an input image can be explained in terms
of 3D objects, the camera pose and the illumination to produce the
reconstructed image. Sometimes a full 3D version may be too onerous,
and we can consider a layered ``2.1D'' model,
where the objects (people) are represented by ``sprite''
models of the shape and appearance, along with the background, as in
Figure \ref{fig:fj}.  The layers have a depth ordering, so that
occlusions can be explained.

Although for simplicity we have limited the scope of this paper
to \emph{static} scene understanding, the structured generative model
approach can also be applied to dynamic scenes. In this case the
objects, camera and lighting can vary (i.e., move, transform etc.)
over time, and one also requires dynamic models of the interactions
between objects.

The goals of this paper are: to promote the SGM viewpoint;
to review relevant work on object
and scene modelling, and inference with SGMs; and to identify
gaps/outstanding issues where further research is needed.
The structure of the paper is as follows: In sec.\
\ref{sec:su_examples} we discuss the rich variety of tasks associated with
scene understanding. Sec.\ \ref{sec:gen_adv} describes the general
advantages of generative models, and sec.\ \ref{sec:pros_cons}
discusses the pros and cons of structured generative
models. Sec.\ \ref{sec:objects} describes modelling objects, including
``things'' (object categories that have a well defined shape) in
sec.\ \ref{sec:things}, including parts-based models, and ``stuff''
(categories which have amorphous spatial extent) in
sec.\ \ref{sec:stuff}. Sec.\ \ref{sec:scenes} covers models of the
inter-relationships of objects, focusing mostly on indoor
scenes. Having defined models for objects and scenes,
sec.\ \ref{sec:inference} discusses how inference for the SGM may be
carried out, from input consisting of a single or multiple
images. Sec.\
\ref{sec:sgm_agenda} discusses issues  that are needed to advance the
SGM agenda, including datasets and benchmarks.

Note that this paper covers a very large amount of ground, and does
not aim to provide comprehensive references for each
topic. Indeed, the necessary topics cover much of the content of
a textbook on computer vision.  Rather, it aims to use prominent
examples of work to provide an illustration of the various topics.

\begin{figure*}[t]
  \begin{tabular}{cc}
    \includegraphics[width=7.7cm]{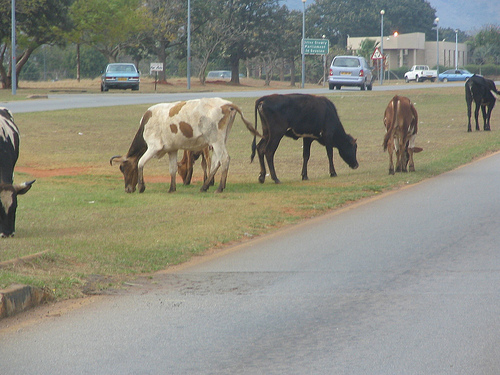} & 
    \includegraphics[width=7.7cm]{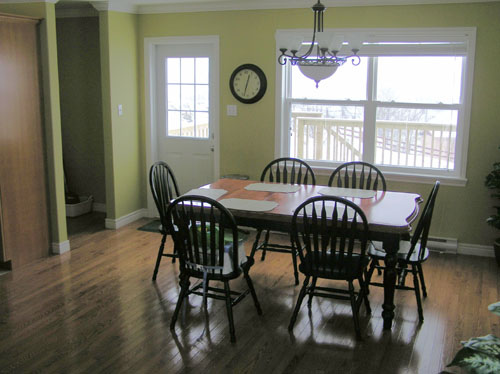} \\
    (a) & (b)
  \end{tabular}    
  \caption{Two example images
    2008\_001062 and  2008\_000043 from the PASCAL VOC 2008 dataset.
 \label{fig:su_examples}}
\end{figure*}

\subsection{Scene Understanding \label{sec:su_examples}}
Before discussing scene understanding more generally, let's first look
at two example images in Fig.\ \ref{fig:su_examples}, and see what we
can extract from them. These are not particularly complex scenes---it
would be easy to pick images with a lot more objects and relations.
Consider first the outdoor scene, Fig.\ \ref{fig:su_examples}(a)---we
can identify that this is not taken
in a dense, ``downtown'' area, but equally not in a very
rural area.  We can identify objects: a small herd of 6 cattle (one is likely a calf
mostly hidden behind the white-and-brown cow near the centre of the
image); 5 motor vehicles (one is half occluded); a building,
some lamp posts; some trees; and some road signs. This focus on
objects may have distracted us from the fact that large amounts
of the images are ``stuff'' categories, such as road surface and grass.
The cows are on the grass, which makes sense as they can graze there,
but not on the road surface.

The second scene, Fig.\ \ref{fig:su_examples}(b), is an indoor scene
of a dining room. We notice a table, 6 chairs (of the same or similar
design), 5 tablemats on the table,
a clock, a light fitting. The room has an
outside door, a window, and a recessed area and a large wooden
cabinet (?) off to the left. There is
a polished wooden floor which allows some reflections, and the walls
are a  yellow-greenish colour. Closer inspection would pick up smaller
details like light switches, a low-level radiator, some objects on the
chairs (including perhaps a child's booster chair), and part of an
indoor plant at the bottom right. We can also see some outdoor
railings through the window, suggesting that there is a stairway up to the area
outside the door.

So what is \emph{scene understanding}? Part of it is about identifying
the objects and the stuff that are visible, but it is more than this.
For each object we would like to know its category, shape and its pose
relative to the scene, and the materials it is made of. We also would
like to understand the camera parameters (e.g., observer viewpoint)
and lighting; this will help explain occlusions and shadows. For
example in Fig.\ \ref{fig:su_examples}(a) one can make inferences for
the direction of the sun, given the shadows of the cattle. Such a 3D
representation allows counterfactual questions, such as predicting how
the image would change if an object was removed or added, or if it was
viewed from a new direction (novel view synthesis, NVS).  It also
enables \emph{interaction} by an agent in the scene, e.g., by
attempting to herd the cattle.

Scene understanding also includes identifying the scene type (e.g.,
dining room), which will give rise to expectations of what objects
should (and should not) be present. And it is about spatial,
functional and semantic relationships between objects.\footnote{See
\url{https://ps.is.mpg.de/research_fields/semantic-scene-understanding}.}
For example in Fig.\ \ref{fig:su_examples}(a) we might find it
surprising that the cattle are not fenced off from the road to
minimize collisions, but this may depend on the norms of the location
where the image was taken.
And in Fig.\ \ref{fig:su_examples}(b) knowledge of dining rooms
  means we would likely expect as many tablemats as chairs to be set
  on the table---in fact close inspection of the image suggests that
  the ``missing'' tablemat is has been placed on the seat of the chair
  on the right hand side.

One rich approach to scene understanding is via answering
questions. In Visual Question Answering (VQA; see,
e.g., \citealt*{antol-vqa-15}), one requires  textual responses
to questions about images. Recent systems like GPT-4 \citep{gpt4-23}
can produce impressive output, although they can make confident mistakes.
But this might not be the best way to
answer certain questions, such as ``which pixels belong to the
black cow near the centre of the image?''.  The PASCAL VOC challenges
\citep{pascal-voc-10}  asked
three questions: For \emph{classification} the question was ``is there
an object of class X in the image?''. For \emph{detection}, the task
was to predict the bounding box of every object of class X in the
image. And for \emph{segmentation} the task was to label each pixel
with one of the known class labels, or background. For the last two,
textual responses are not the most natural way to answer the
questions. Recent systems such as Gemini \citep{gemini-23}
can produce multimodal output (e.g., text and images) that should
help address the limitations of textual VQA. However, note that
such systems use an entangled vectorial latent state, as opposed to a
structured one, with limitations as discussed above.

The epithet ``a picture is worth a thousand words'' also suggests that text
is not the most efficient way to describe a scene, particularly
  given the ambiguities of natural language. Instead, SGMs provide a 
domain-specific language for scenes. OpenAI's 
text-to-image system DALL-E \citep{ramesh-dalle-21} can generate
impressive output in response to prompts like ``a couple of people are
sitting on a wood bench'', or even a quirky prompt like ``a tapir made
of an accordion'' (see Figs.\ 3 and 2(a) in the paper). However, it
has been reported that requesting more than three objects, negation, and
numbers may result in mistakes and object
features may appear on the wrong object
\citep{marcus-davis-aaronson-22}. An issue here is that in
  the paired text and images sourced from the web, the text 
  will likely not be sufficiently informative about the objects and
  their spatial relationships etc. 
In DALL-E, the textual input is compressed into a vectorial
  representation, which is then used to create the image.  DALL-E and
similar systems are thus unlikely to give sufficient control to
graphic artists and animators, who may be broadly happy with the
output of the system, but may wish to make adjustments and edits. In
order to enable this, we argue that one needs object-based
representations and scene models as advocated for above.

\paragraph{Alternative scene representations}
Above we have contrasted an unstructured generative model with a
single vector of latent variables with a SGM. Below we discuss two
other scene representations.

In \emph{semantic scene completion} (SSC), see, e.g.
\citet{song-yu-zeng-chang-savva-funkhouser-17}, the goal is to
predict both the occupancy and object category 
labels\footnote{Including the empty label for free space.} for voxels in the view
frustrum, both on the observed surfaces and in occluded regions.  This
representation shares some of the features of a SGM in that it
describes the geometry of the scene, and is thus amenable to NVS.
However, voxel-based category labels are an impoverished
representation relative to a SGM in that it does not explicitly
represent objects. This could be mostly achieved by running
post-processing on the labelled voxels, although this would not work
for segmenting touching objects of the same class (e.g., two pillows
lying on a bed).  Also the category label provides very little
information about the object appearance.

Modelling a generative distribution over the voxel-based
representation is very challenging, as it bundles together the shape
of individual objects, their relations, and the geometry of the scene
into one data structure. We argue that it is more effective to
\emph{compose} together models of objects, their inter-relationships
and geometry, rather than trying to model this all at the level of
voxel relationships.

The idea of \emph{composing} together conditional generative
models to create finer-grained control of a scene has been
studied. e.g., by \citet{du-etal-23}. They show how to compose
several diffusion models with AND, OR and NOT operators.  The
authors give an example of creating an image of ``a horse on a sandy
beach or a grass plain on a not sunny day''.  This is parsed into
``A horse'' AND (``A sandy beach'' OR ``Grass plains'') AND (NOT
``Sunny''), making use of four conditional diffusion models, for a
horse, a sandy beach, grass plains, and sunny scenes.  This approach
shares some of the desiderata of SGMs in that it is compositional,
although here the composition is carried out \emph{in the image space}.
However, note that there are not explicit representations of the
objects in the scene, so, e.g., this does not give direct control
and editability of the pose, shape and appearance of the horse in
the above example. Also the camera parameters are not available, so
NVS is not available here.

\begin{figure}[t]
\begin{center}  
    \includegraphics[width=3in]{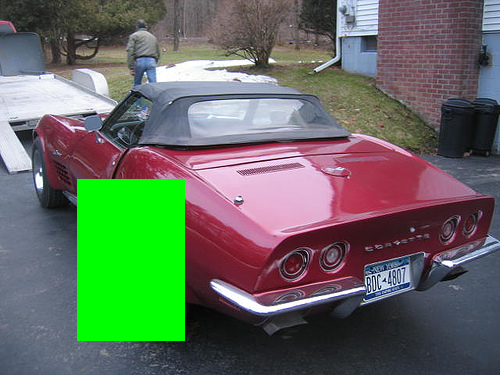}
\end{center}      
  \caption{Image inpainting task, with the green rectangle blanked
  out, based on image 2008\_000959 from the PASCAL VOC 2008 dataset.
 \label{fig:inpainting}}
\end{figure}

\subsection{General Advantages of Generative Models \label{sec:gen_adv}}
We first define some notation: $\bfx$ denotes observed data, such as
an image. A generative model $p_{\theta}(\bfx)$ defines a probability
distribution over images; here $\theta$ denotes the 
parameters of the model. For SGMs, our model is defined in
terms of latent variables $\bfz$.
The latent variables for a single object might be decomposed, for
example into $\bfz = (\bfz^s, \bfz^t, \bfz^p)$, for
shape, texture (including colour) and pose respectively. If there are $K$ objects,
they can each have latent variables (LVs) $\bfz_1, \ldots, \bfz_K$;
let $\bfz_0$ denote the global variables (e.g., camera parameters and
illumination). There can also be additional latent structure in
$\bfz$ that models the
inter-relationships of objects in the scene.  We have that
\begin{equation}
  p_{\theta}(\bfx) = \int p(\bfz) p_{\theta}(\bfx\vert\bfz) \; d \bfz,
\label{eq:xintz}
\end{equation}
where $p(\bfz)$ is a prior over the latent variables, and
$p_{\theta}(\bfx\vert\bfz)$ renders the latent description into the
image.\footnote{In eq.\ \ref{eq:xintz} the integral should be interpreted as
a summation for latent variables that are discrete.}

\sgap

\noi There are several advantages of generative models, as discussed below.

\paragraph{Pattern synthesis (unconditional generation):}
Generative models allow one to sample from the model, and compare the
samples to the raw input data. This can be very helpful, especially to
identify ways in which the model has failed to capture aspects of the
input data, leading to model revision. The methods of \emph{model
criticism} (see e.g., \citealt*{seth-murray-williams-19}) can help
detect differences; one can also train a discriminator to do this, as
used in the training of GANs.

There are situations where such synthetic data (i.e.\ data sampled from
the model) can be useful in its
own right---for example, in healthcare one may not wish to release
data due to privacy concerns. But if one can create synthetic data
which mimics the underlying data distribution, this can enable
research to proceed much more freely. However, one has to be careful
that the synthetic data has not simply ``memorized'' some or all of
the training data, as discussed in \citet{vandenBurg-williams-21}.

\paragraph{Imputation and restoration (conditional generation):}
If the input data $\bfx$ is split into observed data $\bfx^o$ and
missing data $\bfx^m$, then the task of \emph{imputation} is to
predict the missing data given $\bfx^o$.
Probabilistic models also produce a probability distribution for
$p(\bfx^m \vert \bfx^o)$, which allows a quantification of the uncertainty.
In the case that part of an
image is missing, the imputation task can be called \emph{inpainting}; see
Fig.\ \ref{fig:inpainting} for an example. Here a simple method might
just inpaint red and black texture from the car body and tarmac, but
knowledge of cars will predict a wheel in this location, and it is
likely that it will match in style to the visible front wheel.

It might also happen that the observed data is a noisy or degraded
version of the underlying data; in this case having models of the
underlying data and the noise process allows probabilistic restoration of the data.

\paragraph{Anomaly detection:}
It can be helpful to detect datapoints which do not conform to the learned
model $p(\bfx)$. This task is known as \emph{anomaly detection},
\emph{novelty detection} or \emph{out-of-distribution (OOD) detection}.
For example it can be useful for an automated system to detect that
the regime of operation has changed, and thus flag up that it
needs attention or re-training. One way to frame this task is as a
classification between $p(\bfx)$ and a broad (``crud-catcher'') model
$p_0(\bfx)$, as used e.g.\ in \citet{quinn-williams-mcintosh-09}. If a
data point $\bfx$ is more likely under $p_0(\bfx)$, it can be
classified as an outlier relative to $p(\bfx)$.

Anomalies can be quite subtle. In images of street scenes in North
America both fire hydrants and mailboxes are common items of street
furniture, but it is improbable to see a fire hydrant located on top
of a mailbox---this example of a contextual anomaly is from
\citet[Fig.~1]{biederman-mezzanotte-rabinowitz-82}.

\paragraph{Data compression:}
A probabilistic model $p(\bfx)$ can be used to compress
data. Given the true data distribution $p(\bfx)$, 
Shannon's source coding theorem will assign a code of
length $l(\bfx) = - \log_2 p(\bfx)$ bits to $\bfx$. Thus the expected
code length is $- \int p(\bfx) \log_2 p(\bfx) \; d\bfx = H(p)$, the
entropy of $p$. Such data compression can be approached in
practice using, for example, using arithmetic coding, see e.g.\
\citet[sec.\ 6.2]{mackay-03}.

In practice we may not know  
the true distribution $p(\bfx)$, but have an
alternative model $q(\bfx)$. In this case we have to pay a price in terms
of the expected number of bits used. Let the expected code length
when coding under $q(\bfx)$ be denoted $L_q$. Then
\begin{align}
  L_q &= - \int p(\bfx) \log_2 q(\bfx) \; d\bfx  \notag \\ 
  &= - \int p(\bfx) \log_2 \left( \frac{q(\bfx)}{p(\bfx)} 
  p(\bfx) \right) \; d\bfx \notag \\
  &= H(p) + D_{KL}(p \parallel q) \ge H(p) .
\end{align}
Hence the additional expected code length is given by the
Kullback-Leibler (KL) term, which of course reduces to zero when
$q=p$.  This motivates minimizing the KL divergence to produce better
codes, or equivalently to maximize the expected log likelihood $\int
p(\bfx) \log q(\bfx) \; d\bfx$ for a model $q(\bfx)$ (as the entropy
term is fixed).

\subsection{Pros and Cons of Structured Generative Models \label{sec:pros_cons}}
Below we contrast structured generative models (SGMs) compared
to discriminative models, or to unstructured generative models.

\begin{itemize}
\item[\cmark] Structured generative models provide a coherent
  scene representation, rather than just output predictions for a
  disparate set of tasks. This representation is available for
   multiple tasks, including new ones not previously trained on (transfer
   learning).
\item[\cmark] Structured generative models are
  \emph{compositional}. This implies that when learning about a
  particular object type, we don't have to be concerned with other
  object types at the same time. This should make them more data
  efficient.

  If there are inter-object relationships, these can be modelled
  separately from the variability of individual objects, in a
  hierarchical model. This allows
  curriculum learning \citep{bengio-louradour-collobert-weston-09},
  where  one can first focus on modelling individual object classes
  using class-conditional data, and then bring in within-scene
  inter-relationships.
  These advantages are not present in an unstructured generative model.
\item[\cmark] As \citet[sec.\ 7]{yuille-liu-21} argue, vision problems
  suffer from a combinatorial explosion. As we have seen, a single object
  has many degrees of freedom w.r.t.\ object type, pose, shape and
  appearance. These options are then multiplied up by the occurrence
  of multiple objects in a scene.  \emph{Compositionality} assumes that
  scenes are composed hierarchically from more elementary
  substructures, analogous to the ``infinite productivity'' of
  language. By exploiting this compositional structure, SGMs should be
  able to generalize to combinatorial situations without requiring
  exponential amounts of training data.  
\item[\cmark] The SGM representation is \emph{editable}, e.g., to
  change the direction of the lighting, or to add/remove
  objects. In causal language, editing the SGM effects
    \emph{interventions} on the scene representation.
\item[\cmark] (Structured) generative models can be trained
  unsupervised, or with weak supervision. Discriminative models usually
  require potentially expensive/difficult human labelling, although
  the use weaker supervision has also been explored (see, e.g.,
  \citealt*{shi-caesar-ferrari-17}).

A barrier to fully unsupervised learning was highlighted by
\citet{locatello-disentangle-19}, who showed that ``unsupervised
learning of disentangled representations is fundamentally impossible
without inductive biases on both the models and data''. Thus if we
wish to, say, disentangle the shape, texture and pose factors of an
object class, we will need inductive biases in the model. However,
common object models, as described in sec.\ \ref{sec:things}, do in
fact have such inductive biases, so this should not cause problems for
SGMs.
\item[\cmark] The SGM is \emph{interpretable/explainable}.
This structured approach can be contrasted with many
deep generative models, which learn a rich model of the data, but
with a monolithic black-box architecture which is not
interpretable or easily editable. A SGM identifies
certain  image regions as being explained by certain objects,
and can potentially provide more detailed part-level  correspondences.
The structured representation also enables other features such as
occlusion reasoning.
\item[\xmark] Discriminative models can be less susceptible to
  modelling limitations, as they are directly optimizing for a given
  task ``end-to-end'', rather than building a general-purpose model
  which can be used for inference for many different tasks.
\item[\xmark] The SGM framework can require expensive inference
  processes to infer the latent variables for the whole scene. We
  discuss in section \ref{sec:inference} below how these issues can be
  ameliorated.
\end{itemize}  

\begin{figure*}
  \begin{tabular}{cc}
    \includegraphics[width=7.7cm]{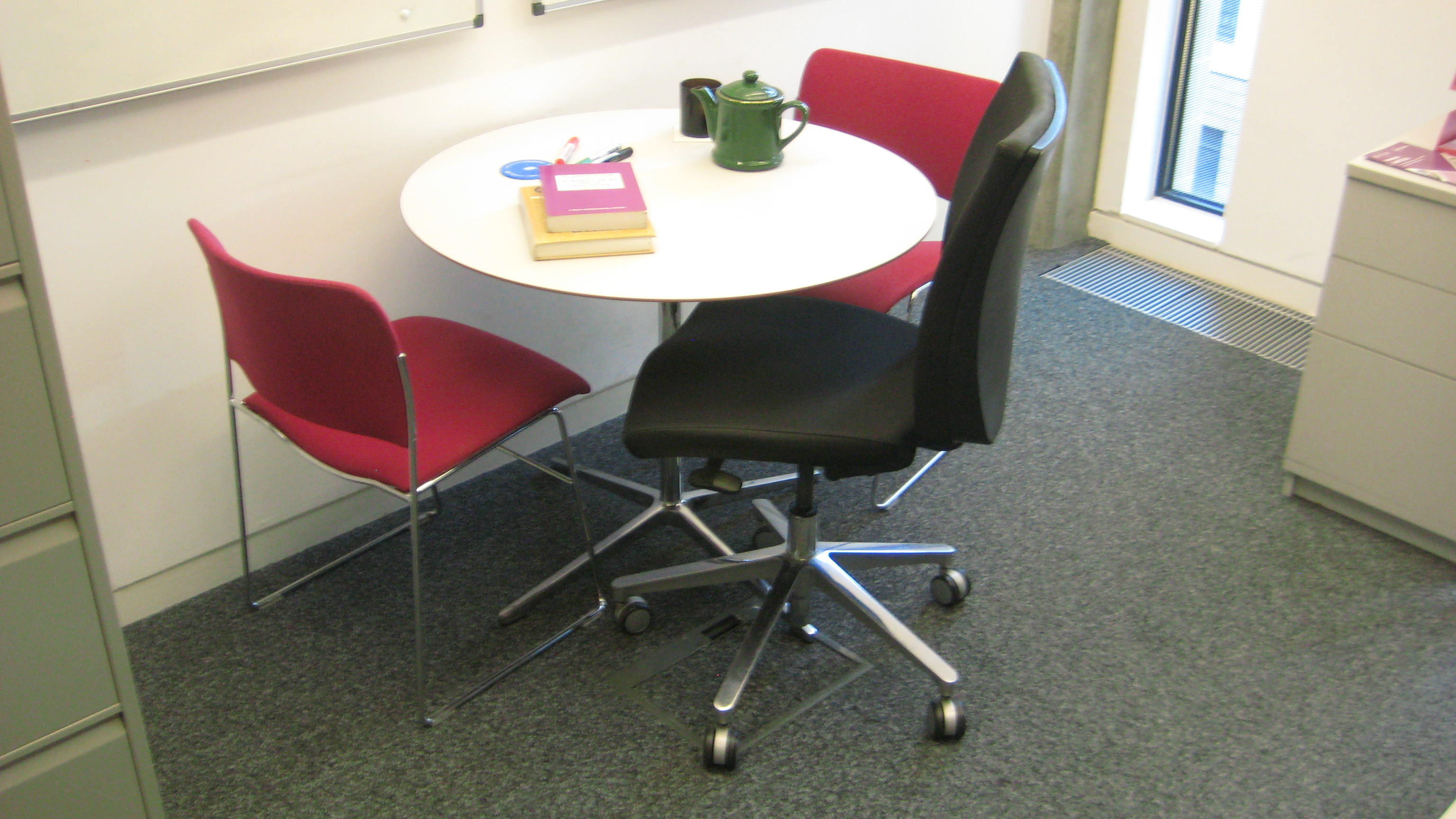} & 
    \includegraphics[width=7.7cm]{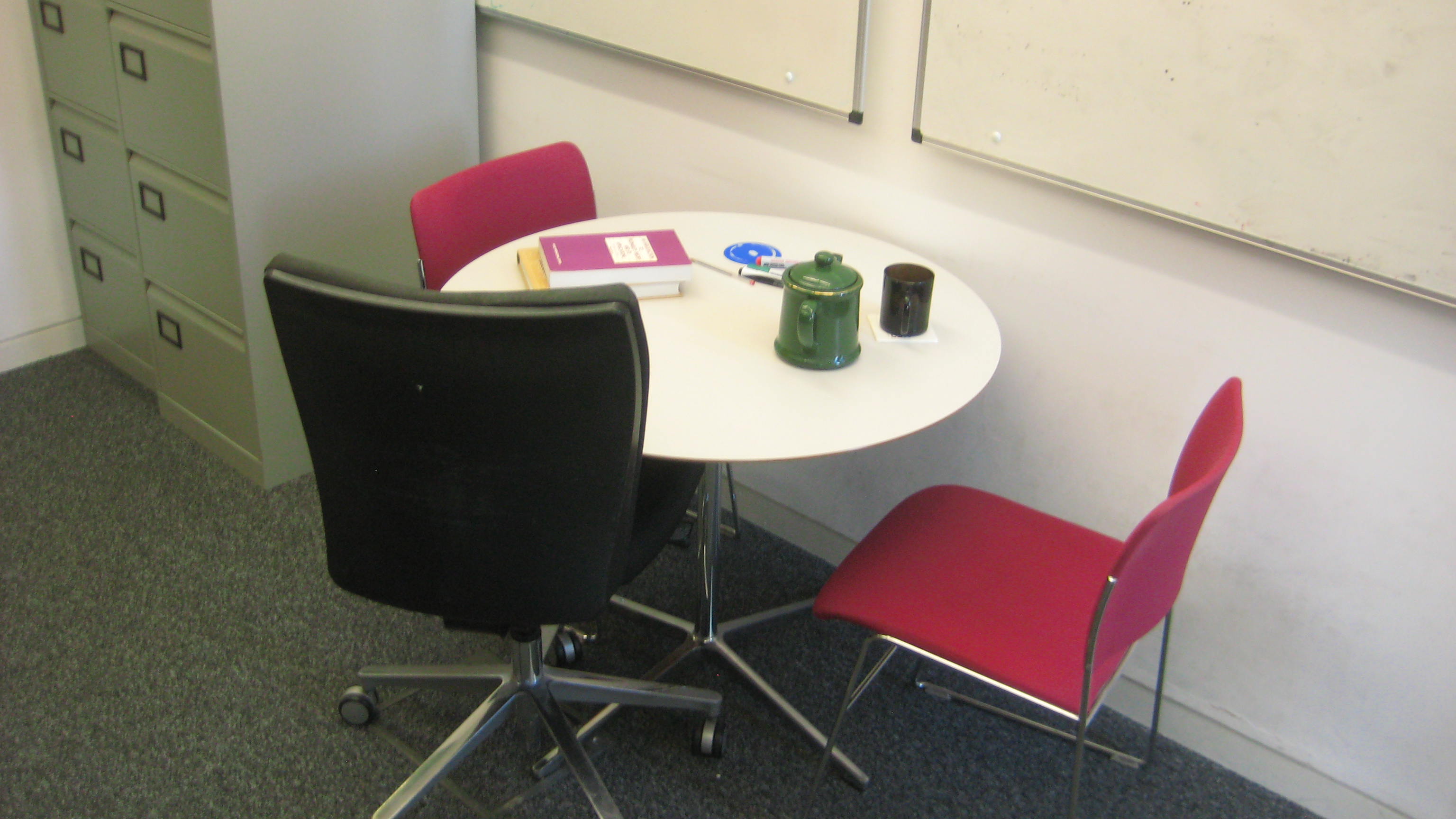} \\
    (a) & (b)
  \end{tabular}    
  \caption{Images of an office scene from two viewpoints. 
 \label{fig:my_office}}
\end{figure*}

An example of where the SGM approach should be helpful is when an
object is heavily occluded, but scene context can help with its
reconstruction. Consider Fig.\ \ref{fig:my_office}(a), where the
rearmost red chair is heavily occluded.\footnote{This example was
inspired by \citet{hueting-reddy-kim-yumer-carr-mitra-18}.}
Knowledge that chairs grouped
around a table are often of the same design in such scenes would help
make strong predictions for this heavily occluded chair.  This could
be evaluated by an instance segmentation of the rearmost chair, where
knowledge of the chair design is likely required to correctly assign
parts of the metal frame visible on the carpet to this chair. 
Other evaluations could include
outputting a 3D model of the object, or making
predictions of how the scene would look from a novel viewpoint, as in
Fig.\ \ref{fig:my_office}(b).  A related task is that of image
inpainting, as illustrated in Fig.\ \ref{fig:inpainting}.  In this
case there is in effect a synthetic occluder (the mask); it is most
natural to evaluate this by prediction of the masked-out region(s).

The SGM approach could also be used to carry out scene editing,
e.g.\ to remove objects or change their properties, or alter the
lighting.  Here the result could be evaluated by collecting views
under the relevant perturbation.  Another possible task is the
completion of a 3D scene given only a subset of the objects in the
scene (see e.g., \citealt*{li-etal-grains-18}). This is a missing data
imputation task, like image inpainting, but different as it is
imputing a 3D scene, not just an image. To achieve this successfully,
the scene model needs to specify new objects of the correct type for
the scene, and to locate them correctly relative to the objects
already present.

\paragraph{Is full inference overkill?} The key advantage of the
SGM approach is that it provides a unified representation, from which
many different tasks or questions can be addressed. This creates a
coherent understanding of the input image in terms of the 3D (or 2.1
D) world, in contrast with what might arise if different models are
trained for different tasks without this underlying structure.

If we only care about one task, then it is usually overkill. But
acting in the real world does not require just one task. The example
of tea-making in \citet{land-mennie-rusted-99} illustrates this
nicely.  The goal of making tea decomposes into subgoals such as
``put the kettle on'', ``make the tea'', ``prepare the cups''.  A
subgoal of putting the kettle on is to ``fill the kettle'', and this
in turn requires ``find the kettle'', ``lift the kettle'', ``remove
the lid'', ``transport to sink'', ``locate and turn on tap'', ``move
kettle to water stream'', ``turn off tap when full'', ``replace lid'',
and ``transport to worktop''. The visual tasks required include
object detection; pose and shape estimation of objects so that they can
be manipulated; and monitoring of the state of some variable
(e.g.\ water level in the kettle). Carrying out the tea-making
task in an unfamiliar kitchen will bring in to play knowledge
about the typical layout of kitchens, and possibly about
different kinds of tap mechanism. Note that subgoals such as
``locate and turn on tap'' and ``turn off the tap'' are
re-usable across other tasks such as making coffee, or
washing the dishes.

\section{Models of Objects \label{sec:objects}}

Visual scenes can contain a lot of complexity.  Components that make
up the scene can be divided into ``things'' and ``stuff''. Things are
object categories that have a well defined shape (like people or
cars), while stuff corresponds to  categories which have an
amorphous spatial extent, such as grass and sky (see e.g.,
\citealt*{sun-kim-kohli-savarese-14}). We first focus on
approaches to model things in sec.\ \ref{sec:things}, and then
move on to model stuff in sec.\ \ref{sec:stuff}.

\subsection{Modelling Things: Multifactor Models \label{sec:things}}
Here we consider modelling a class of visual objects (such as teapots
or cars). These can vary in shape, and in texture (described
e.g.\ by its colour or possibly a more complex pattern on the
surface).  We can also vary the position and
orientation (the pose) of the camera relative to the object, and the
lighting; we term these as \emph{rendering} variables.
Hence there are separate factors of \emph{shape}, \emph{texture} and
\emph{rendering} that combine to produce the observations. We call
models with a number of separate factors {\bf multifactor} models.
Below we first give some examples of multifactor models, and then focus on
parts-based models in sec.\ \ref{sec:parts} (which are a 
special class of multifactor models).

\paragraph{Example: Blanz and Vetter's morphable model of faces.}
 An early example of a 3D multifactor model is due to
 \citet{blanz-vetter-99}.\footnote{The description below is partly
 based on Chapter 17 of \citet{prince-12} as well as the original
 paper.}
Consider $n$ locations on the face; $\bfs_i$ records the
$(x,y,z)$ coordinates of location $i$, and similarly
$\bft_i$ records the red, green and blue colour values (albedo) at the
same location. These measurements were obtained with a laser scanner.
These individual vectors are concatenated to produce the shape vector
$\bfs = (\bfs_1, \bfs_2, \ldots, \bfs_n)$ which has length $3n$,
and similarly there is an texture vector $\bft$ of length $3n$.
\citet{blanz-vetter-99} used approximately 70,000 locations on
each face, and collected data from 200 subjects.
Preprocessing was carried out to remove the
the global 3D transformation between the faces, and an optical
flow method was used to register the locations.

Given the shape and texture vectors for each subject, a probabilistic
principal components analysis (PPCA)
model can be built to capture the variation in shape and texture,
with latent variables $\bfz^s$ and $\bfz^t$ respectively.
One could alternatively use a common $\bfz$ for the shape and texture
variation, e.g.\ by concatenating the $\bfs$ and $\bft$  vectors
for each example before applying PPCA.
This could model the fact that a change in the
shape of the mouth to produce a smile will also likely expose the
teeth to view, so these changes are correlated.

The above description models shape and texture variation in 3D.
This model is transformed geometrically into the image plane in terms
of the camera intrinsic and extrinsic parameters. The colour at
each pixel is determined by the Phong shading model
(see e.g., \citealt*[sec.\ 2.2.2]{szeliski-21}), which accounts for
diffuse and specular reflections from directed light sources, and also
for ambient illumination.
Denoting all of rendering variables by $\bfz^r$, the
overall model can be fitted to a new face by optimizing $\bfz^s$,
$\bfz^t$ and $\bfz^r$ so as to minimize an error measure between the
observed and predicted pixels.

The model of \citet{blanz-vetter-99} is in 3D. Such 3D models have
also been used, e.g., for modelling regions in the brain
\citep{babalola-cootes-twining-petrovic-taylor-08}.  Some earlier work
by Cootes, Taylor and collaborators first developed 2D \emph{active
shape models} using a PPCA model of the shape as defined by landmarks
\citep{cootes-taylor-cooper-graham-95}, and then developed
\emph{active appearance models} that also took the texture into
account \citep{cootes-edwards-taylor-98}.

\begin{figure*}
  \begin{center}
    \includegraphics[height=3in]{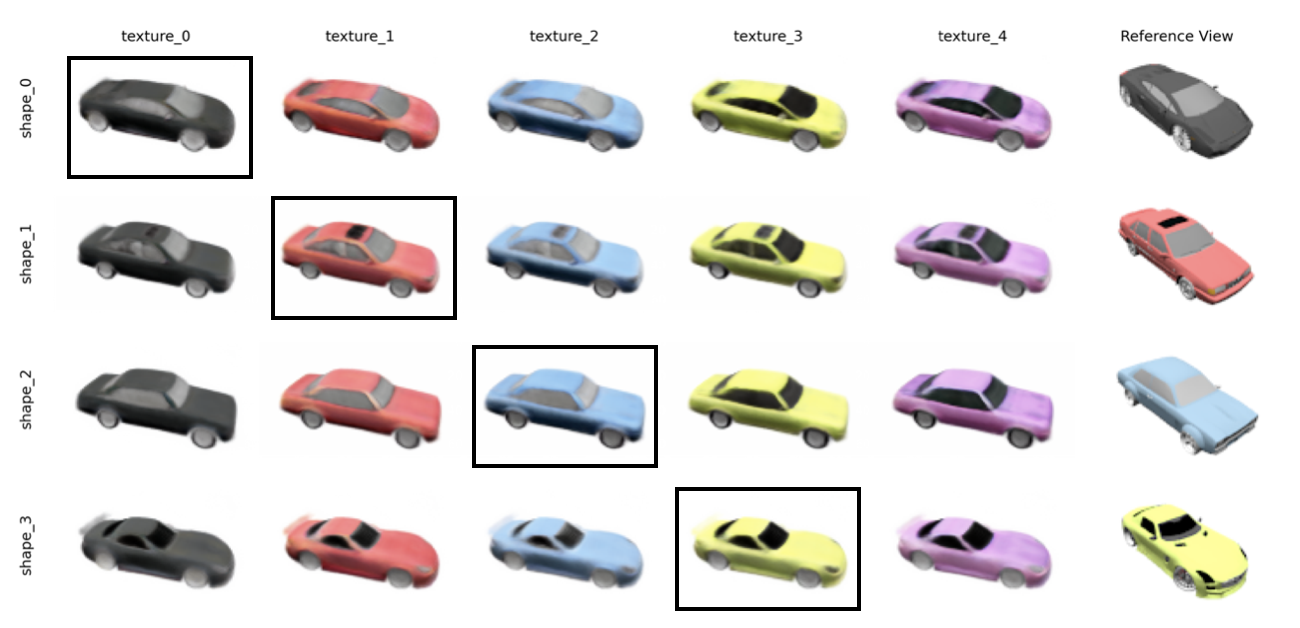}
  \end{center}    
  \caption{An illustration of the ability of CodeNeRF to carry out
    novel shape, texture and pose synthesis. The 4 boxed images
    correspond to renderings with shape and texture codes
    corresponding to the reference views. The other results show
    renders obtained from the cross product of the shape and texture
    codes, at a novel viewpoint.
  Image from Fig.\ 9 in \citet{jang-agapito-21} licenced under CC BY 4.0.
 \label{fig:codenerf}}
\end{figure*}

\paragraph{Example: CodeNeRF models disentangled Neural Radiance Fields
  for object categories.}
Blanz and Vetter's model uses a linear approach (PPCA) to model the
variability due to shape and texture. Careful alignment of the data
was needed in order to make this approach work. With the advent of deep
learning, it is natural to ask if one can exploit more powerful
nonlinear models. We first start with the Neural Radiance Field (NeRF)
representation for a single object due to
\citet{mildenhall-srinivasan-tancik-barron-ramamoorthi-ng-20},
and then add latent variables to model shape and texture variation,
as in the CodeNeRF model of \citet{jang-agapito-21}.

The NeRF takes as input a 3D location $\bfx = (x,y,z)$ and a viewing
direction defined by a 3D Cartesian unit vector $\bfd$, and outputs
a volume density $\sigma(\bfx)$ and
emitted colour $\bfc(\bfx,\bfd) = (r,g,b)$ at that location and direction.
obtained via a neural network $F_{\theta}: (\bfx, \bfd) \rightarrow
(\bfc, \sigma)$ with weights $\theta$. As
\citet[sec.\ 4]{mildenhall-srinivasan-tancik-barron-ramamoorthi-ng-20}
state, ``the volume density $\sigma(\bfx)$ can be interpreted as the
differential probability of a ray terminating at an infinitesimal
particle at location $\bfx$''. This means that the expected colour
$C(\bfr)$  observed at the camera ray $\bfr(t) = \bfo + t \bfd$, with near and
far bounds $t_n$ and $t_f$ is given by\footnote{Note that in this
section $\bfx$ denotes a 3D location, not our usual meaning of an
input image. Also $t$ is overloaded to denote both the parameterization
along the ray $\bfr(t)$, and the texture superscript.}
\begin{equation}
C(\bfr) = \int_{t_n}^{t_f} T(t) \sigma(\bfr(t)) \bfc(\bfr(t),\bfd)
\; dt , 
\end{equation}
where
\begin{equation}
T(t)  = \exp \left( - \int_{t_n}^{t} \sigma(\bfr(s)) \; ds \right) .
\end{equation}
Here $T(t)$ denotes the transmittance along the ray from $t_n$ to $t$,
i.e., the probability that the ray reaches $t$ starting from $t_n$ without hitting any
other particle. The process to obtain $C(\bfr)$ is known as
\emph{volumetric rendering}, and is differtentiable.
The computation of $C(\bfr)$ is approximated by
taking a number of samples along the ray. 

For a single object, the NeRF representation can be obtained by
minimizing the error between the observed and predicted colours at a
set of ray locations for a number of different views (with known
camera poses and intrinsic parameters). This is useful to allow
novel view synthesis, i.e., to predict the image that would be
obtained from a novel view. However, for an object class, it makes
sense to have latent variables $\bfz^s, \bfz^t$ for each object,
as is done in CodeNeRF \citep{jang-agapito-21}.
Here the neural network is enhanced to map
$F_{\theta}: (\bfx, \bfd, \bfz^s, \bfz^t) \rightarrow (\bfc, \sigma)$.
Now the optimization problem for the network weights $\theta$ is carried out
over all training examples in an object class, and for each set of
views  of a given example
the shape and texture latent variables are estimated.
(A regularization penalty proportional to 
$ \vert \bfz^s \vert^2 + \vert \bfz^t \vert^2$ is also imposed on the latent variables,
corresponding to a Gaussian prior.) The ability of CodeNeRF to
generalize to novel shape/texture/pose combinations is illustrated in 
Fig.\ \ref{fig:codenerf}.

There have been a lot of other recent developments arising from the
NeRF work. For example
\citet{zhang-srinivasan-deng-debevec-freeman-barron-21} extended
NeRF to extract a surface (rather than volumetric) representation,
and then solve for spatially-varying reflectance and environment
lighting. This allows rendering of novel views of the object under
arbitrary environment lighting, and editing of the object's material
properties.

\paragraph{Other multifactor models.}
One can sometimes use a \emph{multilinear model} to handle multiple
factors of variation. A multilinear model with two latent factors
$\bfz^1$ and $\bfz^2$ is given by $x_i = \sum_{j,k} w_{ijk} z^{1}_{j}
z^{2}_{k}$, where $w_{ijk}$ is 3-way tensor of
parameters. \citet{tenenbaum-freeman-00} use a bilinear model to 
separate style and content, e.g.\ of the font (style) of different
letters (content).
\citet{vasilescu-terzopoulos-02} use a 4-factor model to
cover different facial geometries (people), expressions, head poses,
and lighting conditions.

Another example of a multifactor model is \emph{transformation
invariant clustering} \citep{frey-jojic-03}.  They consider the case
where there is a discrete factor modelling shape/appearance variation
as a mixture model, and a continuous factor arising from ``nuisance''
translations and rotations of an object in an image.

\subsubsection{Parts-based Models \label{sec:parts}}
Parts-based models are old idea in computer vision. For example
\citet{fischler-elschlager-73} described a model termed 
\emph{pictoral structures}, where an object is represented by
a number of parts arranged in a deformable configuration. The
deformations are represented by spring-like connections between
pairs of parts. \citet{biederman-87} has advocated
for a parts-based approach in computer vision, under the name of
\emph{Recognition-by-Components}.
More recently \citet{felzenszwalb-girshick-mcallester-ramanan-09} used
discriminatively-trained parts-based models to obtain state-of-the-art
results (at the time) for object recognition.

Some advantages of the parts-based approach
are described by \citet{ross-zemel-06}, viz.\
\begin{itemize}
\item A partially-occluded object can be recognized if
  some of the parts are present;
\item A parts-based approach is a good way to model the variability in
  highly-articulated objects such as the human body;
\item Parts may vary less under a change of pose than the appearance
of the whole object.
\item Known parts may be recombined in novel ways to model a new
object class.
\end{itemize}

Parts-based models share a similar objective with \emph{perceptual
organization} or \emph{perceptual grouping} (see, e.g., \citealt*[ch.\
6]{palmer-99}) in that they seek to organize parts into a whole, but
generally perceptual organization is seen as a generic process, e.g.,
for grouping edges into contours, or to group similar pixels into
regions, rather than exploiting specific knowledge about certain object
classes.

\sgap

Below we give four examples of parts-based models.

\begin{figure*}
    \includegraphics[width=\textwidth]{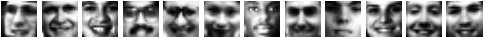}
    \\\\
    \includegraphics[width=\textwidth]{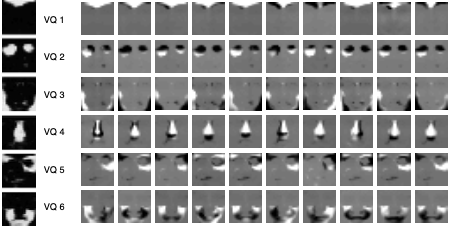} 
  \caption{Top row: sample training images for Ross and Zemel's
    models. Lower rows: parts-based model learned by MCVQ. The plots
    on left show the masks, the probability with which each pixel
    selects the given VQ. On the right are the 10 means for each VQ,
    multiplied by the mask shown on the left. Figures reproduced from
   \citet{ross-zemel-06} with permission of R.\ S. Zemel.
 \label{fig:rosszemel}}
\end{figure*}

\paragraph{Example: Parts-based Models of Faces.}
One common application of parts-based models is to faces. In the UK,
the ``PhotoFit'' system is used by police in the investigation of crimes
to produce an image of a suspect with the help of eye witnesses. See,
for example, the ``PhotoFit Me'' work of Prof.\ Graham Pike.\footnote{See
\url{https://www.open.edu/openlearn/PhotoFitMe}.}. This decomposes a face
into eyes, nose, mouth, jaw and hair parts.

\citet{ross-zemel-06} proposed two models to learn 
such a parts-based decomposition. We describe them in relation to the
modelling face images.\footnote{The data used is
aligned with respect to position and scale, so these ``nuisance
factors'' do not need to be modelled during learning.}
The first model,  Multiple Cause Vector Quantization (MCVQ),
has $K$ multinomial factors,  each selecting
the appearance of a given part. The ``masks'' for each part
are learned probabilistically, so that pixel $j$ is explained by part
$k$ with multinomial probability $\pi_{jk}$, where $\sum_k \pi_{jk} = 1$.
The learned model is illustrated in Figure \ref{fig:rosszemel}. The
plots on the bottom left show the mask probabilities, and on the
bottom right the 10 means for each VQ are shown,
multiplied by the relevant mask. Notice in the bottom left panel
how the masks identify regions such as the eyes, nose and chin.

A second model is termed Multiple Cause Factor Analysis (MCFA)---this
is similar to MCVQ, but now the appearance each part is based on a factor analyzer
instead of a discrete choice. The mask model is as for MCVQ.
\citet{nazabal-tsagkas-williams-23} used a MCFA model of faces, but 
added a higher-level factor
analysis model to correlate the factor analyzers for each part.
The dataset used they used (from PhotoFit Me) is balanced by gender
(female/male) and by race (Black/Asian/Caucasian), hence the
high-level factor analyser can model regularities across the parts,
e.g.\ with respect to  skin tone.

The MCVQ and MCFA models use a simple model for the mask probabilities
$\pi_{jk}$. But suppose we are modelling a parts decomposition of
side-views of cars, e.g.\ into wheels, body and windows. The different
styles of cars will give rise to quite different mask patterns, and
these can be modelled with a latent variable model. For example
\citet{eslami-williams-11} modelled the mask patterns with exponential
family factor analysis, while \citet{eslami-williams-12} used a
multinomial Shape Boltzmann machine.

\begin{figure*}[t]
\centering
\includegraphics[width=5in]{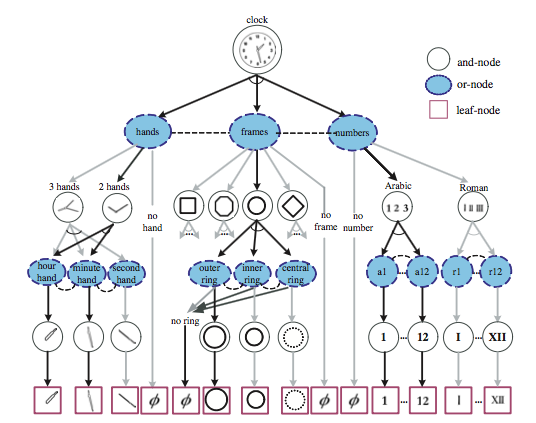}
\caption{AND-OR graph for the clock object category. 
  The dashed links
  between the children of an AND node represent relations
  and constraints. Reproduced from  \citet[Fig.\ 6.1]{zhu-mumford-06}
  with permission of now publishers inc.\ \copyright 2006.
\label{fig:and_or_clock}}
\end{figure*}

\paragraph{Example: Parts-based Model of a Clock.}
Fig.\ \ref{fig:and_or_clock} is reproduced from the work
\citet{zhu-mumford-06}. The figure describes an AND-OR graph for clocks;
for example there is an AND over the hands, frame and numbers
components of the clock, but in each there are alternatives as encoded by OR nodes.
Particular choices at the OR nodes gives rise to a \emph{parse graph}.\footnote{See
sec.\ \ref{sec:hier_scenes} for further discussion of grammars and
AND-OR graphs.}
One of the possible parse graphs is  illustrated with dark arrows,
corresponding to the image at the top. Thus we observe that OR
choices are made for the hands (2 not 3), the numbers (Arabic not
Roman), and the shape of the frame (circular, and only the outer ring
is present).

The AND-OR structure is similar to what we have seen for faces
with the MCVQ; there are a number of parts (the AND), and for each there are
a discrete set of choices (the OR). However, an AND-OR graph is generally more
powerful, as it can have hierarchical structure. For example, in the clock model
there are choices for all of the outer, inner, and central rings
of the frame component to be present or absent.

\paragraph{Example: Parts-based Models of Articulated Bodies.}
A classic example of an articulated object is the human body, which
can be decomposed into a torso, head, arms and legs. The arms
and legs can each be further decomposed; for example an arm is made up
of the upper arm, lower arm and the hand, and the hand can be further
decomposed into the fingers and thumb. Given the importance of
human avatars in the film and gaming industries, there has been
a lot of work on this topic. Here we focus on the Skinned Multi-Person
Linear Model (SMPL) of
\citet{loper-mahmood-romero-pons-moll-black-15}.
There are two factors of variation that need to be taken into
account. The first is the pose of the body, as defined by the
joint angles along the kinematic tree. The second is the
specific body shape of a given individual. The model for the body is defined
by a mesh with some 6890 vertices in 3D.  Each vertex $i$ is
assigned weights $w_{ki}$ indicating how much 
part $k$ affects vertex $i$. 

The body shape is modelled with a linear basis, similar to Blanz and
Vetter's model discussed in sec.\ \ref{sec:things}. The pose of the
body is determined by the axis-angle representation of the relative
rotation of a part with respect to its parent in the kinematic tree.
One other important part of the model is \emph{pose blend shapes},
which modify the the vertex locations depending on the pose, but
\emph{before} the pose transformation is applied. In SMPL, the pose
blend shapes are a linear function of the elements of the part
rotation matrices. Pose blend shapes are needed to counter the
unrealistic deformations at joints that would otherwise arise. After
combining the shape and pose blend effects in the neutral pose, the
final predicted location of vertex $i$ is obtained as a weighted sum
(with weights $w_{ki}$) of the transformation of vertex $i$ under part
$k$.

\paragraph{Example: Capsules} Another parts-based model is termed 
``capsules''. This term was introduced
in \citet{hinton-krizhevsky-wang-11}, with later developments including
\citet{sabour-frosst-hinton-17}, \citet{hinton-sabour-frosst-18} and
\citet{kosiorek-sabour-teh-hinton-19}. There is a recent survey paper
on capsules by \citet{ribeiro-duarte-everett-leontidis-shah-22}.
The term ``capsule'' relates to
a visual entity which outputs both the probability that the
entity is present, and a set of instantiation parameters for the
entity.  Although capsules are usually described in an inferential
manner, with the flow of information from the parts to the object, 
below we follow the exposition of
\citet{nazabal-tsagkas-williams-23} who described
Generative Capsule Models.

Consider an object template $T$ which consists of $N$ parts
$\{ \bfp_n \}_{n=1}^N$. Each part $\bfp_n$ is described by its class,
pose, shape, texture. The template $T$ has an associated latent
variable vector $\bfz$ which affects to the pose, shape, and texture
of the parts. For example for a template in 2D, part of $\bfz$ may
define the parameters of a similarity transformation (in terms of a
translation $\bft$, rotation $\theta$ and scaling $s$ of the
template). The geometric transformation between the object and the
parts can be described by a linear transformation in terms of $\bft$,
$s \cos \theta$ and $s \sin \theta$.  Other parts of $\bfz$ can model
shape and texture correlations between parts.
Methods for matching observed parts to template parts are described in
sec. \ref{sec:inference}.

Although we have described here object-parts relationships, capsules
can be formed into an hierarchical architecture, allowing e.g., the
representation of object inter-relationships.

\subsection{Modelling Stuff: Visual Texture \label{sec:stuff}}

\citet[p.\ 164]{forsyth-ponce-12} discuss texture as follows:
\begin{quotation}
  Texture is a phenomenon that is widespread, easy to recognise, and
hard to define. Typically, whether an effect is referred to as texture
or not depends on the scale at which it is viewed. A leaf that
occupies most of an image is an object, but the foliage of a tree is a
texture. Views of large numbers of small objects are often best
thought of as textures. Examples include grass, foliage, brush,
pebbles, and hair. Many surfaces are marked with orderly patterns that
look like large numbers of small objects. Examples include the spots
of animals such as leopards or cheetahs; the stripes of animals such
as tigers or zebras; the patterns on bark, wood, and skin. Textures
tend to show repetition: (roughly!) the same local patch appears again
and again, though it may be distorted by a viewing transformation.
\end{quotation}

Textures can be classed as regular or stochastic, although there can
be gradations, such as near-regular or near-stochastic. Examples of
regular textures include brickwork, tiled floor patterns, and
wickerwork. Examples of stochastic textures include clouds, wood
grain, and foliage. Below we will focus mainly on stochastic textures.

We take the goal of \emph{texture synthesis} to be the generation of an
arbitrarily sized region of a texture, given a (small) training
sample. For regular textures it should be possible to extract the
repeating element(s) and simply tile the target region appropriately,
but this will not work for stochastic textures. Instead we aim to
learn a generative model of the texture from the training sample.
A common type of model used is an \emph{energy-based model} (EBM).
Let $\bfx_{(k)}$ denote a patch of the image centered at location $k$
(e.g.\ a square patch). The \emph{field of experts} (FoE) energy is defined as
\begin{equation}
  E_{FoE}(\bfx)  = \sum_k \sum_j \phi_j(\bfw_j \cdot \bfx_{(k)}),
  \label{eq:foe}
\end{equation}
where $\bfw_j$ is a filter the same size as the patch, and $\phi_j()$
is some function. A probability distribution over images is then
defined
by the Boltzmann distribution, i.e.,
\begin{equation}
  p_{FoE}(\bfx) = \frac{1}{Z(W)} \exp( - E_{FoE}(\bfx)) .
  \label{eq:boltzmann}
\end{equation}
Here $Z(W)$ is the \emph{partition function} that serves to normalize $p_{FoE}(\bfx)$.
The term  field of experts was introduced in \citet{roth-black-05},
as a generalization of the product of experts construction due to
\citet{hinton-02}, to handle arbitrarily-sized images.

In general it is not easy to draw samples directly from an
energy-based model; a standard approach is to construct a Markov chain
whose equilibrium distribution is the desired Boltzmann distribution. See
e.g., \citet[ch.\ 24]{murphy-12} for further details.

One simple choice would be to take the function $\phi_j$ to be a
quadratic form in $\bfx_{(k)}$. If this is positive definite,
$p_{FoE}(\bfx)$ will be well-defined and normalizable, and will define a
Gaussian Markov random field (GMRF), see e.g.,
\citet[sec.\ B.5]{rasmussen-williams-06}. Stationary GMRFs on a
regular grid can be analyzed via Fourier analysis \citep{rozanov-77}
in terms of the power spectrum. However, it is well known that image
models based on simply on the power spectrum (or equivalently, on
second-order statistics, via the Wiener-Khinchine theorem) are
inadequate. For example, Fig.\ 2 in \citet{galerne-gousseau-morel-11}
shows a section of a tiled roof.
Randomizing the phase of its Fourier transform, while maintaining
the power spectrum, leads to a blurry image, as the phase alignment
needed to create sharp edges no longer occurs.

A Gaussian random field model can also be obtained via the
\emph{maximum entropy principle} (see e.g.,
\citealt*[ch.\ 12]{cover-thomas-91}) on the basis of second order
(covariance) constraints on the distribution. But an alternative is to
consider a set of filters, and impose the constraint that the maximum
entropy model matches the observed histogram for each filter. This
gives rise the FRAME (Filters, Random field, and Maximum Entropy)
model of \citet{zhu-wu-mumford-98}.
\citet{roth-black-05} discuss the FRAME model, and note that
the approach is complicated by its use of discrete filter histograms.
Instead they propose the field of experts model.

\begin{figure*}[t]
\begin{center}
\begin{tabular}{ccccccc}    
  Raw &
  \includegraphics[width=0.13\textwidth]{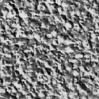} &
  \includegraphics[width=0.13\textwidth]{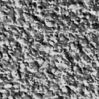} &
  \includegraphics[width=0.13\textwidth]{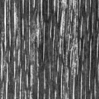} &
  \includegraphics[width=0.13\textwidth]{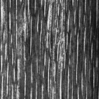} &
  \includegraphics[width=0.13\textwidth]{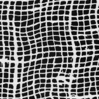} &
  \includegraphics[width=0.13\textwidth]{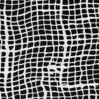} \\
  Tm &
  \includegraphics[width=0.13\textwidth]{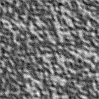} &
  \includegraphics[width=0.13\textwidth]{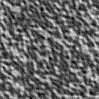} &
  \includegraphics[width=0.13\textwidth]{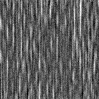} &
  \includegraphics[width=0.13\textwidth]{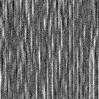} &
  \includegraphics[width=0.13\textwidth]{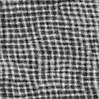} &
  \includegraphics[width=0.13\textwidth]{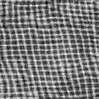} \\
\end{tabular}
\end{center}
\hspace*{35mm} D4 \hspace*{45mm} D68 \hspace*{40mm} D103
\caption{The top row shows two sample patches from each of three textures
  (labelled D4, D68 and D103). The bottom row shows two samples
  drawn from a Tm model trained on the specific textures. Figure
  from \citet[Fig.\ 4]{kivinen-williams-12}.
 \label{fig:TmBrodatz}}
\end{figure*}

\citet{kivinen-williams-12} defined a different energy based model,
with the energy function 
\begin{multline}
  E_{Tm}(\bfx) = \frac{1}{2 \sigma^2} (\bfx - \bfa)^T (\bfx - \bfa) \\  
 -   \sum_{k, j} \log[1 + \exp(b_j + \sigma^{-1} (\bfw_j \cdot
    \bfx_{(k)}) ] . \label{eq:Tm}
\end{multline}  
This was inspired by the convolutional restricted Boltzmann machine of
\citet{lee-grosse-ranganath-ng-09}, but instead of convolution uses
\emph{tiled convolution} as described in
\citet{ranzato-mnih-hinton-10}, who argue
that convolutional weight sharing creates problems due to nearby
latent variables of the same filter being highly correlated. In the
tiled convolutional strategy each filter tiles the image with copies
of itself without overlaps (i.e.\ the stride equals the filter
diameter).  But different filters do overlap with each other, in order
to avoid tiling artifacts.
The energy function $E_{Tm}$ consists of two terms. The first
component corresponds to a
simple spherical Gaussian with mean $\bfa$. The second is obtained
by integrating out the hidden units (as given in 
eq.\ 1 of \citealt*{kivinen-williams-12}) of the restricted  Boltzmann
machine analytically.
The label ``Tm'' is given to this model in order to denote that it
uses tiled convolution, and that the means, rather than the
covariances, are modelled (by the $\{ \bfw_j \}$s and $\bfa$).

\citet{kivinen-williams-12} showed that to model multiple textures one
can keep a fixed set of filters, but adjust the biases on a
per-texture basis---they called this the ``Multi-Tm'' model.
Fig.\ \ref{fig:TmBrodatz} shows results for three different textures,
with two images of both the raw data and samples. The raw data is
obtained as $98 \times 98$ patches cropped
from the Brodatz texture album \citep{brodatz-66}.

An alternative to energy based models is an auto-regressive model,
where the predicted pixel value $x_{ij}$ at location $(i,j)$ depends
on a vector of context, typically to the left and above if working in
raster scan order. \citet{efros-leung-99} used this approach to carry
out \emph{exemplar-based texture synthesis}, starting with a source
training sample of texture. For a target location $(i,j)$, one
identifies neighbourhoods in the source texture that are similar to
the current context region, and then selects one of these regions at
random (depending on the level of agreement with the context region).
For filling in holes an ``onion peeling'' strategy of scanning round
the periphery can be used rather than raster scan order. Note that
this is a non-parametric modelling approach---rather than constructing
a parameterized model for $p(x_{ij} \vert \mathrm{context})$, one extracts
relevant regions from the source texture.

Auto-regressive models do not have to be exemplar-based. PixelCNN
\citep{vandenoord-kalchbrenner-espeholt-kavukcuoglu-vinyals-graves-16} and
PixelCNN+ \citep{salimans-karpathy-chen-kingma-bulatov-17} are
prominent recent auto-regressive models which use convolutional layers for
feature extraction, along
with masking to ensure that only valid context regions are accessed,
in order predict $p(x_{ij} \vert \mathrm{context})$.

Above we have discussed the generation of flat textures on a 2D plane.
Textures can be applied (mapped) to a surface; this is known
as \emph{texture mapping} (see, e.g., \citealt*{szeliski-21}).

\section{Models of Scenes \label{sec:scenes}}

In this section I describe prior work that describes aspects of SGMs.
I first cover ``object-centric models'', which structure the scene
into independent objects, and then discuss models of scenes, based on
autoregressive, energy based and hierarchical models for the
relationships between objects.

A very simple model of scenes is one which randomly selects objects
and puts them into a scene. This might be summarized as ``the
independent components of images are objects''.  One example of this
is the 2D sprites work of \citet{williams-titsias-04} illustrated in
Fig.\ \ref{fig:fj}.  Here the model has learned about the background
and the two people, but a priori it would place the people at random locations in
the image. A more recent example is IODINE (short for Iterative Object
Decomposition Inference NEtwork) due to
\citet{greff-etal-iodine-19}. This uses  an ``object-centric''
representation, consisting of $K$ vectors of latent variables
$\bfz_1, \ldots, \bfz_K$, one for each object.
IODINE was demonstrated on 2D sprites data, and also on images of 3D
scenes from the CLEVR dataset, which consists of geometric objects
like spheres and cubes in random locations with random material
properties. The $\bfz_k$s for each object in IODINE were not factored
into shape, texture and pose components (as discussed in
sec.\ \ref{sec:things}), but were a single vector that entangled these
factors.

However, in the same way that sentences are not random sequences of
words,\footnote{are sequences sentences of not words random!}
visual scenes are not composed of random collections of
objects---there are co-occurrences of objects, and relationships
between them. For example, there are correlations between the scene
type (e.g.\ kitchen. living room, urban street, rural field) and the
kinds of objects observed.  Also, there are stuff-stuff, things-stuff,
and thing-thing interactions that occur between objects in the scene
(see e.g.\ \citealt*{heitz-koller-08}). Examples of things-stuff
interactions are that cars are (usually) found on roads, or cows on
grass. An example
of thing-thing interactions is that dining chairs are likely to be
grouped around a dining table. As another example of thing-thing interactions,
one might consider adding details to a coarse scene
layout, e.g.\ by adding tablemats, cutlery and crockery to the dining table.\footnote{
\citet{ritchie-19} terms this virtual ``set dressing''.}
The reader will observe that there are similarities between
parts-based models described in sec.\ \ref{sec:parts}, and scene
models. However,  parts-based models are often more constrained, with
a fixed number of parts, while scene models can have a variable
number of objects and looser relationships.  Scene relationships
can also be longer-range---a classic example is the relationship
between a TV and a sofa for viewing it, which need to be a comfortable
distance apart.

Below we focus particularly on models for indoor
scenes, reflecting the focus in the research literature.  But
there are also outdoor scenes, in both rural and urban environments.
In scenes of mountains or coastlines, it may be most natural to
consider the carving of the landscape by erosion, e.g.\ by river
valleys. In farmland there will be human-made field boundaries
(constrained by the landscape), along with crops or livestock.  In
urban environments, one might use grammar-type models to generate
building facades, and then ``decorate'' the street architecture with
other objects such as people, cars and street furniture.

Below we describe autoregressive, energy based and 
hierarchical models, which we cover in turn. Note that
autoregressive and energy based models are not latent variable
models, while hierarchical models are.
In this section it is assumed that 3D data is available, e.g.\ 3D
oriented bounding boxes (OBBs) with class labels.

\subsection{Autoregressive Models}
We take as an example the work of 
\citet{ritchie-wang-lin-19}, who describe a process where objects are
added to a room layout one at a time, until a decision is made to
stop. The model first extracts a top-view floor plan of the room (to
define the valid region to place objects). It then feeds the floor
plan to a sequence of four modules that (i) decide which object (if
any) to add, (ii) specify where the object should be located, 
(iii) its orientation, and (iv) its physical dimensions. Once
an object has been added the floor plan representation is updated to
include the object, before the next calls to steps (i) to (iv).
These modules are implemented with convolutional neural networks.

Let the $m$ ordered objects be denoted
$\bfx_1, \; \bfx_2, \ldots, \bfx_m$. Each $\bfx_i$ is comprised
of an object class label, pose features (location and orientation),
shape features and texture
features\footnote{\citet{ritchie-wang-lin-19}
do not use texture features, but in general they could be present.},
so that $\bfx_i = (\bfx_i^c, \bfx_i^p, \bfx_i^s,\bfx_i^t)$. Then
under an autoregressive model we have that
\begin{equation}
  p(\bfx_1, \bfx_2, \dots, \bfx_m) = p(\bfx_1) \prod_{i = 2}^m
  p(\bfx_i \vert \bfx_{<i}),
\end{equation}
where $\bfx_{<i}$ denotes the sequence $\bfx_1, \ldots, \bfx_{i-1}$.
This is suitable for generating scenes from the model. It can also be
used to compute $  p(\bfx_1, \bfx_2, \dots, \bfx_m)$ if an ordering
of the objects is given.
However, when we observe an \emph{unordered} set of objects $X = \{ \bfx_i
\}_{i=1}^m$, we should sum over all possible permutations, so that
\begin{equation}
 p(X) = \sum_{\pi  \in \Pi}  p(\pi) p(\bfx_{\pi(1)}) \prod_{i = 2}^m
  p(\bfx_{\pi(i)} \vert \bfx_{\pi(<i)}),
\label{eq:ARperm}  
\end{equation}
where $\pi$ denotes a particular permutation,and $\Pi$ is the set of all
permutations. The prior over permutations $p(\pi)$ can be taken as
uniform, i.e.\ $1/(m!)$.  In fact \citet{ritchie-wang-lin-19}
do use an ordering of objects for training the object class label
module, based on a measure of the importance of the class.
This depends on the average size of a category multiplied by
its frequency of occurrence.
This means that large objects like a bed will occur first
in a bedroom scene, with other objects fitting in around it.

The model of \citet{ritchie-wang-lin-19}
is relatively simple, but the use of the floor plan
representation with the added objects means that the chain rule of
probabilities can be used readily to model the context, without
simplifications such as a Markov model.\footnote{Although note that
the floor plan representation discards the order in which objects were
added.}
One can ask if the autoregressive process really makes sense as a generative
model of scenes? It is not unreasonable that large objects (such as
the bed in a bedroom scene) should be added first. But it seems
rather unlikely that people furnish bedrooms using a fixed ordering of all
the object classes.

Note also the autoregressive model does not make explicit groupings of
objects that co-occur, such as a bed and nightstand, or a tv-sofa
combination arranged for convenient viewing.  In contrast,
hierarchical models (see below) should be able to pick out such
structure.

Building on the work of \citet{ritchie-wang-lin-19},
\citet{paschalidou-atiss-21} develop an autoregressive model using
transformers (dubbed ATISS).
Rather than summing over all permutations of the sequence as in eq.\
\ref{eq:ARperm}, the authors  randomly permute the order of the objects, and
train the transformer to predict the next object given some previous
ones. In order to handle the fact that a number of different objects
could be predicted, they use mixture models over the object features. 
They demonstrate some advantages of ATISS
over the earlier work, e.g.\ in relation to scene completion,
especially when objects that come early in the sequence (e.g., beds
for a bedroom) are omitted.

\subsection{Energy-based Models}
Suppose that we have generated a number of objects to go in a room.
We then need to arrange them in order to obey a number of spatial,
functional and semantic relationships or constraints.  The basic idea
is to define an energy function that measures the fit of the
configuration to these constraints, and then as per
eq.\ \ref{eq:boltzmann} to define a probability distribution using the
Boltzmann distribution.
\citet{yu-yeung-tang-terzopoulos-chan-osher-11} used such an approach
to automatically optimize furniture arrangements, using simulated
annealing to search in the configuration space.

The method of \citet{yu-yeung-tang-terzopoulos-chan-osher-11} is
defined for a fixed set of objects.
\citet{yeh-yang-watson-goodman-hanrahan-12} extended this idea
to allow an ``open world'', where the number of objects is variable.
They used a reversible jump Markov chain Monte carlo (MCMC) method
from this space.

A problem with energy based models in general is that one generally needs
to run a MCMC chain for many iterations in order to draw samples
from the equilibrium distribution. This limitation applies to
the above methods, meaning EBMs will not scale well to larger scenes.

\subsection{Hierarchical Models \label{sec:hier_scenes}}
A natural way to obtain an hierarchical model is via a {\bf grammar-based
approach}.  The idea of using grammars for pattern analysis is an old
one, see for example the work of K.~S.~\citet{fu-82} on
syntactic pattern recognition. In relation to scene understanding,
perhaps the most notable work is from Song-Chun Zhu and collaborators.
The long paper by \citet{zhu-mumford-06} entitled ``A Stochastic
Grammar of Images'' is a key reference. A parts-based model of clocks due
to \citet{zhu-mumford-06} is discussed above in sec.\ \ref{sec:parts}.

\begin{figure*}[t]
  \centering  
   \includegraphics[width=6in]{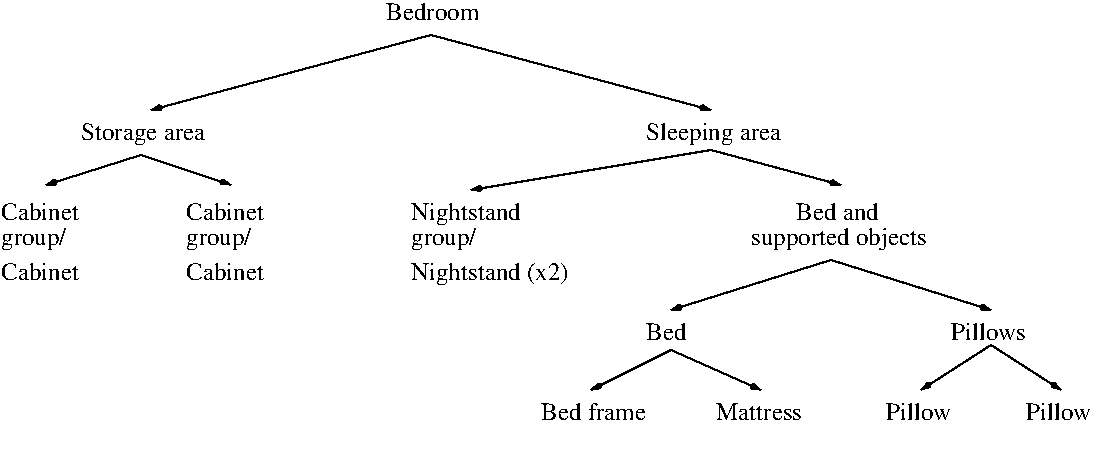} 
  \caption{A hierarchical grammar for bedroom scenes, redrawn from
\citet{liu-chaudhuri-kim-huang-mitra-funkhouser-14}.    
 \label{fig:liu2014Fig4}}
\end{figure*}

A context-free grammar (CFG) is defined in terms of terminal symbols,
non-terminal symbols, a start symbol, and production rules. Inspired by
\citet{liu-chaudhuri-kim-huang-mitra-funkhouser-14}, we take
bedroom scenes as an example domain.
Here we will have terminal symbols for observed objects such as mattress,
nightstand, pillow etc. As above, the full description will
involve not only the class label, but also pose, shape and texture information.
Non-terminal symbols here would
correspond to groupings of objects; for example we may have
$\mathrm{sleeping \mhyphen area} \rightarrow \mathrm{bed} \;
\mathrm{nightstand \mhyphen group}$. There may typically be one or two
nightstands (usually depending on whether the bed is single or
double), so that the productions
$\mathrm{nightstand \mhyphen group} \rightarrow \mathrm{nightstand}$ and
$\mathrm{nightstand \mhyphen group} \rightarrow \mathrm{nightstand} \; \mathrm{nightstand}$
are both valid. An example grammar for bedroom scenes from
\citet{liu-chaudhuri-kim-huang-mitra-funkhouser-14} is shown in Figure
\ref{fig:liu2014Fig4}.

A probabilistic context-free grammar (PCFG) adds a probability
distribution over the productions which share the same non-terminal on
the LHS of the rules. \citet[ch.\ 5]{charniak-93} provides a good overview
of CFGs and PCFGs.
A CFG is often described in terms of a set of production rules, but
it can also be described by an AND-OR graph; 
\citet{hall-73} showed the equivalence between the two. Here, the
OR occurs over productions share the same non-terminal on the LHS;
the AND occurs with productions with more than one symbol on the RHS,
as all of these symbols must be generated.
As \citet[sec.~6.1]{zhu-mumford-06} state, an
``AND-OR graph embeds the whole image grammar and contains all the
valid parse graphs''.

We have seen that the terminal symbols include class label, pose,
shape and texture information in general. This means that the
non-terminals which govern a set of terminals will need to include latent
variables which can generate the appropriate correlations between
them. For example the production 
$\mathrm{sleeping \mhyphen area} \rightarrow \mathrm{bed} \;
\mathrm{nightstand \mhyphen group}$
will need to specify the size of the bed (single or double), and this information
will also need to be passed to the $\mathrm{nightstand \mhyphen group}$ variable,
so that it can determine whether to generate one or two nightstands
(and their appropriate location(s)).
In the CFG, pose and size information can be defined relative to
higher level variables. So, for example,  if the size and location of the
$\mathrm{sleeping \mhyphen area}$ has been defined, it makes sense to locate
the bed relative to this. A possible alternative to encoding this information
in the latent variables is to add horizontal relational structures
that to encode contextual information or constraints between nodes,
as proposed in  \citet[ch.\ 4]{zhu-mumford-06} and \citet{jin-geman-06}.

A problem with using CFGs to model visual scenes is that it can be
hard to learn them from data. Even in natural language processing (NLP),
most successful grammar learning uses annotated ``treebank'' data,
rather than unannotated sentences. Similarly, annotated data has been
used to learn some hierarchical models for images, as in
\citet{yao-yang-zhu-07} and
\citet{liu-chaudhuri-kim-huang-mitra-funkhouser-14}. But recently for
NLP data,
\citet{kim-dyer-rush-19} found that they could add a sentence-level
continuous vector latent variable in addition to the PCFG structure
(to produce a ``compound'' PCFG) yielding state-of-the-art
results for parsing tasks.

While grammars are one way to generate hierarchical structure, they
are not the only way. Instead of discrete-valued non-terminals,
one can use a continuous-valued latent vector in a node.  Consider
starting with a single latent vector $\bfz$. The first binary split can be
generated as
\begin{equation}
[\bfz_{1}, \bfz_2] = \bff(\bfz),
\end{equation}
where $\bfz_{1}$ and  $\bfz_2$ are the latent vectors of the two
children., and $\bff$ is  a
non-linear vector-valued function, which could be as simple
as $[\tanh(W_{1} \bfz + \bfb_{1}), \tanh(W_{2} \bfz
+ \bfb_{2})]$ or a deeper neural network. For each generated
latent vector, a binary node classifier is applied to decide whether it
is a terminal (generating an object), or a non-terminal which can be
further split.  This kind of generative structure was described in
\citet{pollack-90}, and is a forerunner of the
recursive variational autoencoder described below.
The advantage of the continuous state is that it is well
suited to express pose, shape and texture variation.
Note that the hierarchical MCFA model for faces of
\citet{nazabal-tsagkas-williams-23} (described in
sec.\ \ref{sec:parts}) is a simple one-layer model of this type, but with
$K$ child nodes rather than just two, and without the $\tanh$ nonlinearities.

\citet{li-etal-grains-18} developed the Generative
Recursive Autoencoders for INdoor Scenes (GRAINS) model, which
is a recursive variational autoencoder. It takes
as input an unstructured set of objects in 3D and computes a latent
scene representation $\bfz$, which is decoded to yield a hierarchical 
structure of objects which can then be rendered.  In more detail, the
GRAINS model decodes the latent vector $\bfz$ into five variables
representing the floor and four walls of a room. Each of these is
either a terminal (corresponding to an object), or a non-terminal, as
determined by a node classifier neural network. There are number of
non-terminals, corresponding to support, surround and co-occurrence
relationships.
Terminals decode to objects, specified by the OBB, position,
  orientation and semantic label.
A tree-structured encoder is built by first clustering the
objects in the scene to the nearest wall, and then building
sub-trees for each cluster, making use of the non-terminals. A
tree-structured neural network then encodes the scene into the
latent $\bfz$.

A more recent recursive variational autoencoder model is the Scene
Hierarchical Graph Network (SceneHGN) due to
\citet{gao-sun-mo-lai-guibas-yang-23}. This does away with the
dependence on the four walls, and builds a hierarchical scene
representation with levels of the room, functional regions, objects
and object parts. It also uses a different set of non-terminals.

It should be noted that the encoder structure used in recursive VAEs
is not required to mirror exactly the decoder structure; it only needs
to output the latent vector $\bfz$.  The heuristic tree construction
used in GRAINS and SceneHGN can be contrasted with a grammar, which
sums over all parse trees. This construction can considerably reduce
the complexity of inference compared to the full search/summation over
trees.

\section{Inference \label{sec:inference}}

In this section the task is inference from a single RGB or RGBD image,
or from a set of images with known viewpoints (multi-view data),
assuming that the object and relational models have already been learned.
Thus we are interested in $p(\bfz \vert \bfx)$.

We first cover inference for objects and global variables, and
then inference with scene models. Finally we discuss inference
in the presence of model deficiency, where an input image cannot
be reconstructed exactly.

\paragraph{Inference for Objects and Global Variables.}
The basic inference task is to (i) detect each object, and to determine
its class label, shape, texture and pose information, (ii) detect
and characterize regions of stuff, and (iii)
determine the lighting and camera global latent variables. 
The inference problem for a single object is already daunting if we we
consider a discretization of the values for each of the latent
variables and search over these (see e.g.\ \citealt*{yuille-liu-21}
sec.\ 7.1), and there is a combinatorial explosion when considering
multiple objects and the global variables.  The most direct way to
address the combinatorial explosion is to \emph{approximate} the
search.  One such method is Markov chain Monte Carlo (MCMC), where
the goal is construct a Markov chain whose equilibrium distribution
samples from the desired posterior over the latent variables. 
See e.g., \citet[chapter 24]{murphy-12} for a discussion of MCMC.
Often the proposed moves in the state-space are generic and do not
exploit the structure of the problem, leading to slow mixing. An example of a more
advanced approach is Data-driven Markov chain Monte Carlo (DDMCMC,
\citealt*{tu-zhu-02}), which allows proposals such as the candidate set of
object detections to be incorporated into a valid MCMC algorithm.

An alternative approximate approach is to use \emph{variational
inference}, see e.g.\ \citet{jordan-ghahramani-jaakkola-saul-99},
where the goal is approximate the posterior $p(\bfz\vert\bfx)$ with
a variational distribution $q(\bfz)$. The variational
autoencoder \citep{kingma-welling-14,rezende-mohamed-wierstra-14} uses
\emph{amortized} variational inference to predict a distribution over
the latent variables given input data. This is achieved with 
an ``encoder'' network (a.k.a.\ a ``recognition model'', see
\citealt*{dayan-hinton-neal-zemel-95}).
Such amortized inference is relatively straightforward if
there is one object of interest in the image, but is more complex if
there are multiple objects, due e.g.\ to permutation symmetries.
In IODINE \citep{greff-etal-iodine-19} the feed-forward predictions
from the image for the object latent variables are the refined iteratively
to take into account effects such as explaining away. In the
slot-attention model of
\citet{locatello-etal-20}, an iterative attention mechanism is used
in the mapping from the inputs to the latent variables, so that they
\emph{compete} to explain the objects in the image. The
attend, infer, repeat (AIR) model of \citet{eslami-air-16} takes an
alternative, sequential approach, identifying one object at a time.

MCMC and variational inference (VI) methods can be used to express the
uncertainty in the latent representation $\bfz$ given the data
$\bfx$. This may arise, e.g.\ due to (partial) occlusions, and can give
rise to a multi-modal posterior, corresponding to different
interpretations or ``parses'' of the input image.  A limitation of VI
methods is that they may not capture this multi-modality well if the
assumed form of $q(\bfz)$ is unimodal.  As explained in
\citet[sec. 21.2.2]{murphy-12}, when the variational distribution is
unimodal, then the Kullback-Leibler divergence $KL(q \! \parallel \! p)$ used in VI
tends to pick out one mode of the posterior, and thus under-estimate
the uncertainty.  MCMC methods can in principle sample from a
multi-modal posterior, but in practice can get trapped within one
mode, unless special measures such as annealing (see e.g.,
\citealt*[sec. 24.6]{murphy-12}) are used.
One way to reduce the uncertainty in $q(\bfz)$ is to fuse
information from multiple views, as shown
in \citet{nanbo-eastwood-fisher-20}.

The above models such as IODINE and AIR were trained unsupervised. 
But the development of object detectors over the last two
decades means that one can train object detectors for known classes
such as cars, pineapples etc. The output may be a bounding box (BBox), or
possibly a region-of-interest. Such an approach was used by
\citet{izadinia-shan-seitz-17} in their IM2CAD system, which takes
as input an image of a room. It estimates the room geometry (as a 3D
box), detects objects, predicts the associated latent variables of
each object, places the objects in the scene, and then optimizes their
placement via rendering the scene and comparing it to the input image.

If the input data is a set of multi-view images, then geometric
computer vision techniques can be brought to bear. Simultaneous
localization and mapping (SLAM) or structure-from-motion (SfM) can be
used to estimate camera poses and a sparse point cloud representation,
as in COLMAP \citep{schoenberger-frahm-16}.  By themselves, such
techniques do not segment the data into objects, but they can be
combined with object-detection bounding boxes and masks predicted from
each image to produce a 3D bounding box, as in \citet[sec.\
5.1]{ruenz-etal-FroDO-20}. Once the objects are segmented, their
shape, texture and pose latent variables can be estimated, see e.g.,
\citet[secs.\ 5.2-5.3]{ruenz-etal-FroDO-20}.

One important recent development has been \emph{differentiable
rendering}, which allows optimization of the estimated latent state,
in order to improve the fit between rendered scene and the input
image(s). An early differentiable renderer was \emph{OpenDR} due to
\citet{loper-black-14}. \citet{kato-diff-render-20} provide a
survey of various methods that have been proposed for mesh,
voxel, point cloud and implicit surface representations.

\paragraph{Inference with Scene Models.}
Above we have discussed instantiating objects and the camera/lighting
global variables; we now consider inference for
\emph{relational} structures.
An autoregressive transformer model like ATISS \citep{paschalidou-atiss-21}
can compute the joint probability of $M$ objects under its
model in $O(M^2)$ time (see, e.g., \citealt*[sec.\ 22.4]{murphy-23}),
The random permutations used in training
effectively allow it to model a \emph{set}, rather than a sequence.

Things are more complex for hierarchical models, where a parse graph
may need to be constructed on the fly.  One of the major issues is to
match a set of object and part detections to a scene structure like an
AND-OR graph. As a first example, consider a capsules model for an
object, as discussed in sec.\ \ref{sec:parts}. This is comprised of a
number of parts that lie in a certain geometric relationship to each
other.  (Similar arguments also apply to a grouping a objects in a
scene, such as a computer-monitor-keyboard, but here we will use the
terminology of an object and its parts.)

Now consider that we have detected a set of parts
$\{ \bfx_m \}_{m=1}^M$, and the task is to match these to a set of
templates $\{T_k \}_{k=1}^K$. Let $w_{mnk} \in \{ 0, 1 \}$ indicate
whether observation $\bfx_m$ is matched to part $n$ of template $k$.
The $w$'s can be considered as a binary matrix $W$ indexed by $m$ and
$n,k$.  One way to frame this assignment problem is to consider
valid \emph{matchings} between observed and template parts, as defined
by a permutation matrix. If $M$ is not equal to the total number of
model parts $N = \sum_{k=1}^K N_k$, then dummy rows or columns can be
added to make the problem square, e.g., in case some parts are
missing.  As the exact computation would require considering all
possible permutations which scales exponentially with $M$,
\citet{nazabal-tsagkas-williams-23} consider variational inference
for the match variables $W$ and the latent variables $\bfz_k$ for each
template. The complexity of this is $O(M^2)$ per iteration, arising
from computing real-valued variables $r_{mnk}$ that approximate the
posterior probabilities of $W$ under a mean-field assumption that
respects that the row and column probabilities  sum to one.
This implements a \emph{routing-by-agreement} procedure, as
discussed by \citet{sabour-frosst-hinton-17} and
\citet{hinton-sabour-frosst-18}, but derived from
the variational inference equations rather than as an {\it ad hoc}
objective.  An alternative to routing-by-agreement is to use a random
sample consensus approach (RANSAC, \citealt*{fischler-bolles-81}),
where a minimal number of parts are used in order to instantiate an
object template, which is then verified by finding the remaining parts
in the predicted locations.
\citet{kosiorek-sabour-teh-hinton-19}
use another inference approach, where
an autoencoder architecture predicts the LVs for each template, making
use of a Set Transformer architecture
\citep{lee-lee-kosiorek-choi-teh-19} to handle the arbitrary number of
observations $M$.

For hierarchical models, algorithms for image parsing
are discussed, for example, in \citet[sec.\ 8]{zhu-mumford-06} for
AND-OR trees, and
in \citet{liu-chaudhuri-kim-huang-mitra-funkhouser-14} for PCFGs. In
sequences (such as text) the complexity of inference for a PCFG model
is $O(M^3)$ for a sentence of length $M$, as the inside-outside
algorithm can be exploited (see e.g., \citealt*[ch.\ 6]{charniak-93}).
However, without this sequential ordering property, exact inference is
more expensive, and approximations can be used.

A recent development for inference is GFlowNets, see e.g.,
\citet{hu-malkin-jain-evertt-graikos-bengio-23}. Here the idea is to
train a stochastic policy to sample from a target distribution over a
set of objects ${\cal Z}$ (such as complete parse trees). This is
achieved by sampling from a larger set of objects ${\cal S}$ of
partially constructed objects (like incomplete parse trees). These
objects are built up sequentially from an initial empty state until a
valid object in the target distribution is obtained, via a forward
policy $p_F(s' \vert s, \bfx)$. The training objective learns $p_F$ so as to
produce valid samples from the target distribution, here $p(\bfz\vert\bfx)$.

\sgap

If the hierarchical structure has been defined in terms of a
recursive neural network, as in GRAINS \citep{li-etal-grains-18}, then
it is possible to build an encoder to predict the scene latent variable $\bfz$,
which can then be decoded to produce the hierarchical structure. This
was successful in GRAINS, but note that there the input is a set of
segmented 3D objects, as opposed to an unsegmented image.

\sgap

Although it is natural to think of inferential information flows in the
hierarchical model being bottom-up, from parts to objects to scenes,
it does not have to happen this way. The overall
scene type may be well-characterized by a global scene descriptor like
the \emph{gist} \citep{oliva-torralba-06}, and this will create, e.g.,
top-down expectations for certain object classes, and not others.
For example, \citet{torralba-murphy-freeman-10} made use of the gist
to predict the scene type (such as beach scene, street scene, living room). This
then made useful predictions for the vertical location in the image
for objects of a certain class (e.g.\ cars in street
scenes)\footnote{In the scene types considered, the horizontal
location of objects was usually not well constrained by the scene type.}.
This indicates more generally that information from the image(s)
may flow in top-down as well as bottom-up fashion for inference 
in the scene model.

As \citet[sec.\ 1.3.4]{zhu-huang-21} note, vision is driven by a
large number of tasks, and ``Each of these tasks utilizes only a
small portion of the parse graph, which is constructed in real-time in
a task-driven way based on a hierarchy of relevant tasks''.  Thus
it may be useful to consider ``lazy
inference''\footnote{Thanks for Kevin Murphy for suggesting this term.}
of the full parse graph.

\paragraph{Model deficiency:} Quoting George Box, we can say that ``all
models are wrong, but some are useful'' \citep[p.\ 424]{box-draper-87}.
There is a great richness and
detail in many visual scenes, and it may not be possible (or 
perhaps even desirable) to model this fully. However, if we cannot get zero
error at the pixels, to what extent can a generative model be said
to have fully explained the image? The issue here is that
there can be many ways in which an aggregate measure of error (such as
mean squared error, MSE) can arise: (i) the pose of an object could be
slightly off, but
the appearance and shape are correct, leading to a ``halo'' of errors
around the boundaries of the object; (ii) the pose and shape may be correct,
but the object's texture does not match the palette of known textures,
and thus there is a high-frequency pattern of errors within the
object's extent; (iii) an object's shape is incorrect (either
due to failures in inference or modelling), but the pose and texture
are correct; or (iv) there may have been a false positive or false
negative detection of a small object, again leading to the same MSE.

A natural way to tackle this problem is to compare the input and
predicted images, along with the predictions for object extent etc.
To my knowledge there has only been a little prior work on prediction
of the quality of the outputs of a vision algorithm when ground truth
is not available.
\citet{jammalamadaka-zisserman-eichner-ferrari-jawahar-12} discuss
\emph{evaluator algorithms} to predict if the output of a human pose
estimation algorithm is correct or not, and 
\cite{xin-zhang-liu-shen-yuille-20} have looked at predicting failures
in semantic segmentation.  In the reconstructive framework the
goodness-of-fit of the geometric variables (camera parameters, object
pose and shape) can be measured by the intersection-over-union (IoU)
of the the predicted and ground-truth object masks (as used e.g.\ in
\citealt*{romaszko-williams-winn-20}).  Thus this IoU measure could be
predicted by an evaluator algorithm. Errors in these variables will
have consequent effects on the pixel errors. But if the IoU is
satisfactory, then object-level pixel errors will likely be due to
errors in the texture or lighting variables. Assessing the
significance of such errors with an evaluator algorithm
will require the annotations of the severity and types of the errors
made.

It may be thought that algorithms that provide estimates of the
\emph{uncertainty} in their predictions help to address the issue of the
assessment of vision algorithms, and this is partially correct.
However, a limitation of probabilistic model uncertainty is that it is
assessed \emph{relative} to a model (or a fixed set of models). But
if, for example, the model's palette of textures does not match with
that in an input image, then the model's posterior uncertainty
measures will not characterize the true situation well. This
deficiency is known as the ``open world'' (M-open) situation, in
contrast to M-closed, as discussed e.g.\ by \citet[\S
6.1.2]{bernardo-smith-94}. One approach to addressing this is
via \emph{model criticism}, as discussed e.g., by
\citet{seth-murray-williams-19}.

Given the complexity of visual scenes, it may not be possible all the
detail in a scene at the pixel level. Instead, one can build a
generative model of the spatial layout of \emph{image features}, as
computed e.g.\ by a DNN. Such an approach is used, for example, in the
neural mesh model (NeMo) of \citet{wang-kortylewski-yuille-21}.  While
it is more interpretable to reconstruct the input image than a
feature-based representation, the latter may make modelling easier.
The experimental results for the NeMo generative model show that it is
much more robust to OOD tasks like recognition under partial occlusion
and prediction of previously unseen poses than standard feedforward
DNNs.

\section{Advancing the SGM agenda \label{sec:sgm_agenda}}

Above I have laid out the key topics of modelling objects and scenes,
and carrying out inference in these models, and relevant
issues. Below I comment further on the state-of-the-art, and issues
around datasets etc.

\paragraph{Modelling objects:}
The state-of-the-art seems to be at a good level. In terms of data,
the Amazon Berkeley Objects (ABO) dataset of \citet{collins-ABO-22} is a recent
example of a reasonably large collection of models (some 8,000 objects)
with complex geometries and high-resolution, physically-based
materials that allow for photorealistic rendering.

\paragraph{Modelling scenes:} Compared to objects, the
state-of-the-art is less advanced. Much of the focus has been on
indoor scenes, leaving the modelling of outdoor scenes in urban and
rural environments less explored. The task of modelling scenes is also
more difficult than for objects, involving variable numbers of objects
and types, and spatial, functional and semantic relationships between
them.  There is still much to be done here, e.g., exploring the
  pros and cons of autoregressive and hierarchical models, and
  developing new ones. One notable  challenge is to automatically learn
  hierarchical structure from data.
  For hierarchical models, recursive VAEs seem particularly promising,
  as they can exploit the power of neural networks and distributed
  representations, and the flexibility over the choice of the encoder
  architecture, while decoding to an object-based structure.

One issue here is data. Progress in 2D
image recognition has been driven by large-scale datasets. Currently
there are very few large 3D datasets available, and none on the scale
of, say, the ImageNet which contains over one million images.
A notable dataset here is
CAD-Estate \citep{maninis-popov-niessner-ferrari-23} which is derived
from 19,512 videos of complex real estate scenes. Each object is
annotated with  with a CAD model from a database, and placed in
the 3D coordinate frame of the scene. Over 100k objects are so
annotated.
Of course the effort needed to obtain and annotate a 3D scene is much
greater than simply annotating bounding boxes in images.
Computer graphics can, of course, provide a rich source of 3D ground
truth and annotations (see e.g.\ the CLEVR dataset of
\citealt*{johnson2017clevr}), and can provide a controlled means to
test compositional generalization (see their sec.\ 4.7).  However,
this requires good object and particularly scene models to generate
realistic scene
layouts, and also there are issues on how well models trained on such
rendered data will transfer to real scenes.

While collecting 3D datasets is challenging, it is also possible to
collect \emph{multi-view data}, where there are multiple images of
(parts of a) given scene, with known camera parameters (intrinsic and
extrinsic). See, e.g., the active vision dataset from UNC\footnote{
\url{https://www.cs.unc.edu/~ammirato/active_vision_dataset_website/}.}
and the Aria
dataset\footnote{\url{https://about.meta.com/realitylabs/projectaria/datasets/}.}
from Meta. SGMs can be tested on multi--view by predicting what will be
observed from a novel test viewpoint.

\paragraph{Inference:} Scene-level inference is
challenging, with the need to match portions of a hierarchical scene
model with image data. Exact inference for grammars and AND-OR
graphs can scale poorly, but approximations such as  GFlowNets
\citep{hu-malkin-jain-evertt-graikos-bengio-23} and alternative models
such as recursive variational autoencoders may hold promise.
As we have seen above, inference can involve bottom-up
and top-down flows of information in the model. This could lead to
complex inference processes, reminiscent of those found in earlier vision models,
such as VISIONS \citep{hanson-riseman-visions-78} and the system of
\citet{ohta-kanade-sakai-78}. It may be possible to simplify this
somewhat by using the idea of ``lazy inference'', where only parts of
the whole scene representation are activated in response to given
task. With regard to model deficiency, to make progress it will be
important to get lots of data on the kinds of deficiencies that occur most,
in order to address the important issues.

\paragraph{Tasks and Benchmarks:} Currently most tasks for computer vision
are evaluated in the image plane. Partly this may be due to the fact
that it is relatively easy to collect images and create annotations in
this case.  However, there are some 3D benchmarks such as the KITTI
suite\footnote{\url{http://www.cvlibs.net/datasets/kitti/}.} and
nuScenes\footnote{\url{https://www.nuscenes.org}.}. These
include 3D object detection in road scenes (using 3D bounding
boxes). Autonomous driving and bin picking tasks
in cluttered scenes are examples of areas that may well push
the 3D reconstructive agenda forward.

In order to exploit the value of structured generative models, it will
be necessary to define a set of tasks which can exploit the same
underlying representation.  Some examples of challenging tasks were
given in sec.\ \ref{sec:pros_cons}; these include object
reconstruction under heavy occlusion, scene editing, and scene
completion tasks, all of which exploit the 3D and scene-level
information contained in the SGM representation.  The research
community will need to focus attention on a tasteful choice of
3D/multi-view benchmarks and tasks in order to promote the SGM
approach.

\section*{Acknowledgements}
I thank David Hogg, Alan Yuille, Oisin Mac Aodha, Antonio Vergari,
Siddharth N., Kevin Murphy, Paul Henderson, Adam Kortylewski, Hakan
Bilen, Titas Anciukevicius, Nikolay Malkin and Bradley Love
for helpful comments and discussions, and the anonymous referees for
their comments which helped improve the paper.

For the purpose of open access, the author has applied a Creative
Commons Attribution (CC BY) licence to any Author Accepted Manuscript
version arising from this submission.


\end{document}